\pdfoutput=1
\documentclass[11pt]{article}

\usepackage[]{acl}

\usepackage{times}
\usepackage{latexsym}

\usepackage[T1]{fontenc}
\usepackage[utf8]{inputenc}

\usepackage{microtype}

\usepackage{graphicx}
\usepackage{natbib} % has a nice set of citation styles and commands
\usepackage{mathtools} % amsmath with fixes and additions
\usepackage{booktabs} % commands to create good-looking tables
\usepackage{tikz} % nice language for creating drawings and diagrams
\usetikzlibrary{positioning}

\usepackage{multicol}
\usepackage{amssymb}
\usepackage{enumitem}
\usepackage{xcolor}
\usepackage{bbm}
\usepackage{subcaption}
\usepackage{makecell}
\usepackage[labelfont=bf, format=plain,singlelinecheck=false]{caption}
\usepackage{capt-of}
\usepackage[export]{adjustbox}
\usepackage{colortbl}
\usepackage[disable]{todonotes}
\newcolumntype{$}{>{\global\let\currentrowstyle\relax}}
\newcolumntype{^}{>{\currentrowstyle}}
\newcommand{\rowstyle}[1]{\gdef\currentrowstyle{#1}%
  #1\ignorespaces
}

\newtheorem{definition}{Definition}
\newtheorem{example}{Example}

\title{APPTeK: Agent-Based Predicate Prediction in Temporal Knowledge Graphs}

\author{Christian M.M. Frey \\
  Christian-Albrecht University of Kiel\\
  Christian-Albrechts-Platz 4 \\
  24118 Kiel \\
  \texttt{\small cfr@informatik.uni-kiel.de} \\\And
  Yunpu Ma \\
  LMU Munich \\
  Oettingenstr. 67 \\
  80538 Munich \\
  \texttt{\small ma@dbs.ifi.lmu.de} \\\And
  Matthias Schubert \\
  LMU Munich \\
  Oettingenstr. 67 \\
  80538 Munich \\
  \texttt{\small schubert@dbs.ifi.lmu.de} \\}

\begin{document}
\maketitle

\begin{abstract}
%Living in a constantly changing environment, we have to adapt our choices according to the current times we live in. 
%In knowledge graph reasoning, we observe a trend to analyze temporal data evolving over time. 
In \emph{temporal Knowledge Graphs} (tKGs), the temporal dimension is attached to facts in a knowledge base resulting in quadruples between entities such as \emph{(Nintendo, released, Super Mario, Sep-13-1985)}, where the predicate %between two entities is defined
holds within a time interval or at a timestamp.
%Multi-hop reasoning on inferred subgraphs connecting entities within a knowledge graph can be formulated as a reinforcement learning task where the agent sequentially performs inference upon the explored subgraph. 
%The task in this work is to infer the predicate between a subject and an object entity, i.e., \emph{(subject, ?, object, time)}, being valid at a certain timestamp or time interval. 
\todo{shortened abstract}
%Our reinforcement learning agent gathers temporal relevant information about the query entities' neighborhoods, simultaneously. 
We propose a reinforcement learning agent gathering temporal relevant information about the query entities' neighborhoods, simultaneously.
We refer to the encodings of the explored graph structures as \emph{fingerprints} which are used as input to a Q-network.
%Subsequently, we use the two fingerprints as input to a Q-Network. 
Our agent decides sequentially which relation type needs to be explored next to expand the local subgraphs of the query entities. Our evaluation shows that the proposed method yields competitive results compared to state-of-the-art embedding algorithms for tKGs, and we additionally gain information about the relevant structures between subjects and objects.% This provides an additional explainability component of the system's inference.% A thorough evaluation of benchmark datasets \emph{ICEWS14}, \emph{ICEWS05-15}, and \emph{YAGO15K} shows the value of reinforcement learning on tKGs.
\end{abstract}

\section{Introduction}
\label{sec:introduction}
\todo{commented out}
%Nowadays, discrete information is often organized in the form of knowledge graphs (KGs) describing relations between a subject and an object entity. For example, the triple \emph{(Nintendo, released, Super Mario)} yields \emph{Nintendo} as subject, \emph{Super Mario} as object and \emph{released} as predicate. 
%
%In most cases, KGs are automatically extracted from text or web sources as sets of triples. Thus, 
\noindent%
%Most KGs are incomplete in the sense that not all relations holding between its entities are represented by a triple in the KG. 
The task of \emph{Knowledge Graph Completion} (KGC) is about adding missing facts, and a plethora of methods have been proposed for this task \cite{Wang-2014-proppr,lao-etal-2011-random,lin-2018-multihop,Shen2018MWalkLT,das-etal-2017-chains,xiong-etal-2017-deeppath,chen-etal-2018-variational}. In general, there are two ways to complete a knowledge graph. The \emph{entity prediction task} aims at retrieving all entities w.r.t a query predicate. %displaying a particular query relation to a given query entity
%, i.e., queries have the form \emph{(subject, predicate, ?)}. 
In contrast, the \emph{predicate prediction task} is: given two entities, predict the type of relationship holding between these entities\cite{onuki-2019-jour,Lingbing-2019,teru-2020-pmlr}.
%Thus, the corresponding query is \emph{(subject, ?, object)} 

A property of most knowledge bases is that facts change over time. For example, the triples \emph{(Barack Obama, is\_president\_of, USA)} and \emph{(Donald Trump, is\_president\_of, USA)} both held in the past but are now invalid. To be able to represent changing facts over time correctly, KGs have been extended to \emph{temporal Knowledge Graphs} (tKGs). In particular, each fact in a tKG is complemented by the time the fact was valid. %Depending on the type of knowledge, times are either particular timestamps or time intervals. 
In recent research, it has been shown that methods for static KGs are not always feasible for tKGs \cite{Leblay-ttranse-2018,garcia-duran-etal-2018-learning,dasgupta-etal-2018-hyte}, and thus, new methods are needed for completion.

In this paper, we propose a novel technique for \emph{predicate prediction} on tKGs, i.e., queries have the form \emph{(subject, ?, object, timestamp)} . The idea of our method is to train a reinforcement learning (RL) agent that actively gathers information about the relation of subject and object at a given time. This information is represented as a so-called \emph{topology} and technically corresponds to a subgraph of the tKG. Topologies can be used in a downstream task as input to a classifier for predicting the relation types holding between entities. Furthermore, in contrast to embedding models yielding low-dimensional vector representations, topologies can be visually analyzed to give human users an insight into how two entities are connected.

Though there exist several methods for completing static knowledge graphs based on RL \cite{das-2017-minerva,lin-2018-multihop,Shen2018MWalkLT,fu-2019-collaborative}, our method shows significant differences. %First of all, % previous methods did not integrate temporal constraints into the agent and thus, are not directly applicable to tKGs. Furthermore,
First, previous RL approaches rely on finding paths between subject and object. In contrast, our agent is based on topologies that generalizes in the sense that several paths between entities can be found at once. %allows the agent to search for information that cannot be found on a single path between subjects and objects. 
The agent's training procedure employs Q-Learning where encodings (\emph{fingerprints}) of the explored topologies are used as input.
Furthermore, our agent employs a bidirectional search that simultaneously gathers information about both query entities. Finally, as we focus on predicate prediction instead of entity prediction, we only need to train a single agent for all relation types, whereas previous RL methods need to train a dedicated agent for each relation type \cite{xiong-etal-2017-deeppath}.
We compare our method, called \emph{APPTeK}, with KGC techniques for tKGs on three open benchmark datasets in our experiments and demonstrate its benefits. The contributions of this paper are:
\setitemize{noitemsep}
\begin{itemize}
    \item A predicate prediction method for tKGs that is based on topologies instead of paths.
    \item An RL agent that is trained to find these topologies for a given pair of query entities. 
    \item An empirical evaluation of the advantages of our approach on three benchmark tKGs.
\end{itemize}

\section{Related Work}
\label{sec:related_work}
We provide an overview of related works in the field of i) \emph{Reasoning on Static Knowledge Graphs}, and ii) \emph{Reasoning on Temporal Knowledge Graphs}.\\

\noindent% 
\todo{commented out; focus on RL approaches}
i) %In \cite{lao-etal-2011-random}, the authors used random walks with restart to derive paths and applied supervised training to rank those paths (PRA: Path Ranking Algorithm). %This work was generalized by \cite{Wang-2014-proppr} with recursive probabilistic logic programs where other relations are jointly used to infer a target relation. 
%In 2017, \cite{das-etal-2017-chains} introduced the application of recurrent neural networks (RNNs) to perform reasoning of the target relation where the input was given by deriving paths via PRA (Path Ranking Algorithm \cite{lao-etal-2011-random}. 
%
%Multi-hop reasoning has also been formulated as a sequential decision problem allowing the use of reinforcement learning (RL) to perform effective path search. 
In \emph{DeepPath} \cite{xiong-etal-2017-deeppath}, the authors merged ideas from path-based and embedding-based reasoning with deep RL. %. In particular, the learning procedure is based on RL techniques instead of using random walks with restart. 
In \cite{chen-etal-2018-variational}, the authors introduced \emph{DIVA} that bridges path finding and reasoning with variational inference. %\emph{DIVA} infers latent paths connecting nodes within a KG by parameterizing the likelihood (path reasoning) and prior (path finding) with neural network modules.  
For \emph{fact prediction}, \emph{MINERVA} \cite{das-2017-minerva} takes path walking to the correct answer entity as a sequential optimization problem by maximizing the expected reward. %The model excludes the target answer entity and provides a more capable inference. 
In order to tackle the sparsity reward problem, the authors of \cite{lin-2018-multihop} propose a soft reward mechanism instead of using a binary reward function. \todo{check}% Action dropout is adopted to mask some outgoing edges during training to enable more effective path exploration. 
In \emph{M-Walk} \cite{Shen2018MWalkLT}, the model applies a RNN controller to capture the historical trajectory and uses \emph{Monte Carlo Tree Search} (MCTS) for effective path generation.
\todo{commented out; rather NLP}
% By leveraging a text corpus with a sentence bag of current entities, CPL \cite{fu-2019-collaborative} proposes collaborative policy learning for path-finding and fact extraction from textual data. 
MemoryPath \cite{Li-2021-MemoryPathAD} has been proposed as another RL-motivated approach leveraging LSTMs and graph attention mechanisms to form memory components. 

All of the proposed RL approaches for static KGs aim at the entity prediction task and are not applicable to predicate prediction.
\todo{commented out;}
%For static KGs, the predicate prediction task is the main focus in \cite{onuki-2019-jour,Lingbing-2019,teru-2020-pmlr}. GraIL \cite{teru-2020-pmlr} proposed a GNN framework that reasons over local subgraph structures. Lingbing et al. \cite{Lingbing-2019} designed a multi-layer RNN to model triples in a KG as sequences. In this work, the relation prediction task is interpreted with only one given entity, and therefore, this approach is different from our task, where both entities are incorporated into the input data. 
%
%To conclude, all of the named approaches have been introduced for static KGs - therefore, are disregarding the temporal aspect.
Additionally, all of the methods listed above disregard the temporal aspect.

% %%%%
%In contrast to our work, most of the approaches do not specifically handle queries in the form \emph{(subject, ?, object, timestamp)}. Most of the related works handle queries for the \emph{link prediction} task, i.e., queries of \emph{(subject, relation, ?)}, resp., with a timestamp for tKG, where the agents receive information about the relation as input. %Additionally, all of them have been introduced for static knowledge graphs - therefore, disregarding the temporal aspect.
%For example, DeepPath\cite{xiong-etal-2017-deeppath} %includes the answer entity in its inference step, and has been introduced for static KGs. It 
%does not handle larger (temporal) graphs at once as for each query relation an own subgraph is processed. In order to extend DeepPath to tKGs, a proper choice of embedding method is required as initialization, as well as finding a way to incorporate the temporal aspect into path inference and adapting the weighted reward function accordingly.\\
%the temporal aspect would need to be incorporated into the supervised pre-training step first.\hfill\\

ii) Reasoning on tKGs are categorized into interpolation and extrapolation methods \cite{jin2020Renet}. Whereas the former tries to infer missing historical facts \cite{wu-2020-temp,Xu-2020,jung-2021,bai-2021-tpath,xu-etal-2021-temporal}, the latter tries to predict facts that will hold in the future \cite{Han2020graph,jin2020Renet,zhu-2021-cygnet,deng-2020,sun-2021-timetraveler,li-2021-search,bai-2021-tpmod}. 
CyGNet \cite{zhu-2021-cygnet} and RE-NET \cite{jin2020Renet} proposed methods for the entity prediction task by encoding historical facts related to the subject entity into the query. %CyGNet uses a generate-copy network to model repetitive patterns. %, i.e, it captures the frequency of historical facts with the same subject entities and relations as the given queries. 
%In contrast, RE-NET uses a GCN and GRU to model the sequence of $1$-hop subgraphs related to the given subject entity. 
In \cite{li-2021-regcn}, an RE-GCN embedding model is proposed that incorporates an evolution unit, enabling both, entity and predicate prediction. %, i.e., a relation-aware GCN is leveraged to capture the structural dependencies within the KG at each timestamp being able to solve entity and predicate prediction. 
%Generally, embedding models for tKGs incorporate temporal information in their embeddings. 
\todo{check if info necessary}
%They have been proven to have better performances on entity prediction over tKGs, i.e, for answering queries in the form \emph{(subject, predicate, ?, timestamp)}, compared to traditional embedding models on static KGs. Nevertheless, 
Some approaches are built upon formerly introduced heuristics on static KGs, like \emph{TTransE} \cite{Leblay-ttranse-2018}, \emph{TA-TransE} \cite{garcia-duran-etal-2018-learning}, and \emph{HyTE} \cite{dasgupta-etal-2018-hyte}, which are based on the translation-based method \emph{TransE} \cite{bordes-2013-transe}. %In the literature, we also have extensions of the well-known 
Exentions of \emph{DistMult} \cite{yang-2015-embedding} are defined by the models like \emph{TA-DistMult} proposed in \cite{garcia-duran-etal-2018-learning} or Know-Evolve \cite{trivedi-2017-knowevolve}. Also, the \emph{SimplE} model \cite{kazemi-2018-simple} has been extended in \cite{goel2020diachronic} to \emph{DE-SimplE}, which incorporates a dyachronic transformation function for including the temporal information in the latent embedding space. The same technique is also applied on \emph{TransE} \cite{bordes-2013-transe} and \emph{DistMult} \cite{yang-2015-embedding}, resulting in the models \emph{DE-TransE}, resp., \emph{DE-DistMult}.

Even though most of these methods have been introduced to solve the \emph{entity prediction} task, we adapt several methods which currently represent the state-of-the-art in tKG completion to \emph{predicate prediction}. Thus, we can compare our new agent-based approach to the most promising methods for KGC in temporal KGs. Details about the modifications are described in more detail in Sec. \ref{sec:evaluation}.
%An overview of RL-motivated approaches is provided in table \ref{tab:scoring_fncs}.
\section{APPTeK - tKGR Agent}
\label{sec:mira}
\iffalse
In this section, we give a formal problem definition of relation prediction in \ref{sec:problem_definition}. The procedure of how the agent gathers and processes information is presented in \ref{sec:mdp}, and the computation of Q-values is presented in detail in section \ref{sec:model_architecture}.
\fi

%%% SECTION PRELIMINARIES
\subsection{Preliminaries}
\label{sec:problem_definition}
%First, we introduce basic concepts and notations related to the tKG reasoning problem and provide well-established formal definitions.

A \emph{temporal Knowledge Graph} (tKG) can be represented by a set of time-dependent facts $\mathcal{G} = \{(u, r, v, t) | u, v \in \mathcal{E}, r \in \mathcal{R}, t \in \mathcal{T}\}$, where $\mathcal{E}$ is a set of entities, $\mathcal{R}$ is a set of relations and $\mathcal{T}$ denotes the temporal domain. The components of a quadruple are denoted by a \emph{subject} and an \emph{object} entity $u, v \in \mathcal{E}$ being linked by a semantic \emph{predicate} $r \in \mathcal{R}$ at a specific \emph{time} $t \in \mathcal{T}$. The latter can either describe a particular point in time or a time interval.

Next, we define the \emph{(interpolated) predicate prediction} task in \emph{temporal Knowledge Graph Reasoning} (tKGR) for knowledge graph completion (KGC). 

\begin{definition}[(Interp.) Predicate Pred. in tKG]
Given a tKG and a query quadruple $(u_q, ?, v_q, t_q)$, where $?$ is the unknown predicate, in a \emph{(interpolated) predicate prediction problem}%(a.k.a. relation reasoning)
, we try to infer the relation types $r_q \in \mathcal{R}$ holding between $u_q$ and $v_q$ at time $t_q$ given the already known quadruples in the tKG. 
\end{definition}

%In RL on static KGs, the result set for solving \emph{link predictions} relies on queries of the form $(s, p, ?)$ which in recent works \cite{xiong-etal-2017-deeppath,Shen2018MWalkLT,das-2017-minerva} imply direct paths from the subject $s$ to any target object $o$.
Intuitively, we present a model that allows the agent to gather information about $u_q$ and $v_q$, simultaneously. Therefore, the agent's decision is based on inferred subgraphs being constructed by starting from both query entities at once. 
\todo{shortened}
%, whereas models in recent works \cite{xiong-etal-2017-deeppath,Shen2018MWalkLT,das-2017-minerva,sun-2021-timetraveler,bai-2021-tpath} imply single paths from a subject $sub$ to any target object $obj$. Our approach finds relevant information to predict the predicates holding between both entities at the specified query time. 
As the gathered information does not include all information of the $k$-th order proximity of an entity $x$ and is not restricted to paths between both entities, we will refer to the time-dependent observed subgraph for an entity $x$ as the \emph{temporal topology} of $x$, or \emph{topology} of $x$ for short.

% \textcolor{red}{
% For downstream tasks such as predicate prediction, topologies are transformed to binary features which can be used in ordinary classifiers. In our evaluation, we used a softmaxed linear layer  for classification.
% }
%For the downstream task of predicate prediction, the explored topologies are used as binary features that are fed into a single layer classifier. 
%Each topology indicated whether two entities are connected by expanding according to its information or not. 
%In our approach, the relation types stored in a topology are used as logical formulas to derive sparse vector descriptions which can be fed into a classifier to predict relation types.

%%%

%%%%%%%%%%%%%%%%%%%
% MDP 
%%%%%%%%%%%%%%%%%%%
\subsection{MDP Formulation}
\label{sec:mdp}
%The goal of our tKGR agent is to learn how to reason based on known quadruples within the tKG. 
\todo{shortened}
%Our single tKGR agent learns to build topologies containing relevant information for describing the predicate between the source entity $sub$ and target entity $obj$ at time $t$. Specifically, 
%Our agent extends the set of known facts by selecting relation types being adjacent to any of its known entities. 
\todo{check again}
%The set of observed facts within both topologies starting from $sub$ and $obj$ acts as evidence for the prediction task and yields an explanation for the predicted predicate. 
%Hence, our agent learns to build topologies containing relevant information for describing the predicate between $sub$ and $obj$ at time $t$. 
%For the sake of brevity, we will use $x$ to substitute $\{sub,obj\}$ but emphasize that the single agent queries topologies always from both entities, simultaneously. 

%First, we provide the definition of a Markov Decision Process (MDP) and how it applies to our setting.
\noindent%
In the following, we define the Markov Decision Process (MDP) for our setting. An MDP is defined by the tuple $(S, A, R, P)$, where $S$ is the set of states, $A$ is the set of actions, $R$ is the reward function and $P$ is the state transition function. 

For the sake of brevity, we will use $x$ to substitute $\{u_q, v_q\}$ but emphasize that the single agent queries topologies always from both entities, simultaneously.

%%%%%%%%%%%%%%%%%%%
% STATE SPACE 
%%%%%%%%%%%%%%%%%%%
\subsubsection{State Space $S$}
\label{subsubsec:mira_statespace}%\hfill\\
The state comprises the \emph{topologies} visited by the agent, i.e., the known part of the graph. Likewise to traditional graph expansion techniques, we distinguish between already visited entities and entities that can be explored next. We refer to them as the \emph{core} and \emph{periphery} information. 

%Compared to other related work where a re-selection of linkages might be allowed, we force our agent to explore solely parts of the graph that have not been explored yet. 
Considering the query entities $x\in \{u_q, v_q\}$, we denote the core information of a topology as $c_x^i$ and the periphery information as $p_x^i$ in step $i$.

% In order to prevent the agent to re-select relations, we distinguish between sets of relations that have already been taken into account %by the agent 
% and relations that can be explored in succeeding steps. Therefore, we introduce the core information $c_x^i$ %in the $i$-th iteration 
% and the periphery information $p_x^i$ of a topology, starting from the query entities $x\in \{s,o\}$.

%\begin{itemize}
%\item 
\textbf{Core Information.}~The core $c_x^i$ comprises information about quadruples which have been visited by the agent. Technically, it is defined as a subgraph of the input tKG, i.e, $c_x^i \subseteq \mathcal{G}.$ %in preceding steps $i=1,\ldots, i-1$.
%relation types with their respective temporal information which has been considered by the agent in preceding steps $i=1,\ldots, i-1$. 
%Hence, $c_x^i$ consists of the history of all traversed facts. 
% Initially, $c_x^0$ is empty. The information stored in $c_x^i$ is extended step by step with quadruples
% %relations having a timestamps 
% matching the agent's selected relation type. We define:
% \todo{in transition}
% \begin{equation}
%     \label{eq:core}
%     c_x^i \coloneqq \left\{c_{x}^{i-1} \cup y^{i}\right\}  \qquad c_x^0 \coloneqq \varnothing
% \end{equation}
% where $y^i$ denotes the subset of quadruples %\emph{(relation, timestamp)}-pairs, 
% for which the relation type matches the agent's selected action $a^i$. %, which we elaborate on in the next section.
%
%\item 

%%%%%%%%%%%%%%%%%%%%%%%%%
% PERIPHERY INFORMATION %
%%%%%%%%%%%%%%%%%%%%%%%%%
\textbf{Periphery Information.~}The periphery information $p_x^i$ %indicates \emph{(relation, timestamp)}-pairs the agent can select in succeeding steps. 
consists of quadruples $(u_p,r_p,v_p,t_p)$ where $u_p$ is found in $c^i_x$ but $v_p$ is not. In addition, we also have to consider the temporal proximity of the fact time $t_p$ to the query time $t_q$.
%The relations within the periphery information are incident to parts of the graph having already been traversed. 
%Therefore, the herein stored information describes the agent's action space. Details %about the action space 
%are described in the paragraph \emph{Action Space}. 

Formally, the periphery is defined as follows: Let $\mathcal{G} = \{(u, r, v, t) | u, v \in \mathcal{E}, r \in \mathcal{R}, t \in \mathcal{T}\}$ be a tKG and let $c^i_x$ be the core information at step $i$, then the periphery $p_x^i \subset \mathcal{G}$ consists of the following tuples:

% \begin{alignat}{2}
% \label{apptek:eq:periphery}
% p_x^i = \{&(\hat{u},\hat{a},\hat{v},\hat{t}) \in \mathcal{G}\setminus c_x^i |&&\\
% &\exists (\bar u,\bar a, \bar v, \bar t) \in c_x^i: &&(\hat{u} = \bar u \vee \hat{u} = \bar v)~\wedge \nonumber\\
% % \not \exists  (\bar u,\bar a, \bar v, \bar t) \in c_x^i:
% % ((\hat{v} = \bar u \vee \hat{v} = \bar v))\\
% & &&|t_q - \hat{t}| < \Delta tknn\} \nonumber \,,
% \end{alignat}

\begin{alignat}{2}
\label{apptek:eq:periphery}
p_x^i &= \{(u_p, r_p, v_p, t_p) \in &&\mathcal{G}\setminus c_x^i |\\
&\exists ( u_c, r_c, v_c, t_c) \in c_x^i: &&(u_p = u_c \vee u_p = v_v)~\wedge \nonumber\\
% \not \exists  (\bar u,\bar a, \bar v, \bar t) \in c_x^i:
% ((\hat{v} = \bar u \vee \hat{v} = \bar v))\\
& &&|t_q - t_p| < \Delta tknn\} \nonumber \,,
\end{alignat}

where $\Delta tknn$ is the $k-$nearest neighbor distance w.r.t. the query time $t_q$.

We define the state $s_x^{i}$ of each topology starting from $x\in \{u_q, v_q\}$ at iteration $i$ by joining the core (Eq. \ref{apptek:eq:core}) and periphery (Eq. \ref{apptek:eq:periphery}) information:
\vspace{-0.3cm}
 \begin{equation}
    s_{x}^{i} \coloneqq c_{x}^{i} \cup p_x^{i}
 \end{equation}
Finally, the overall state is defined as the union of both topologies. We define: 
 \begin{equation}
     S^i= \bigcup_{x \in \{u_q,v_q\}} s_{x}^{i}
 \end{equation}

% \begin{minipage}{0.4\columnwidth}
% \end{minipage}
% \begin{minipage}{0.59\columnwidth}
% \end{minipage}
% \vspace{0.1 cm}
% \begin{multicols}{2}
% %\noindent
%  \begin{equation}
%     s_{x}^{i} \coloneqq c_{x}^{i} \cup p_x^{i}
%  \end{equation}
%  \begin{equation}
%     S^i= \cup_{x \in \{sub,obj\}} s_{x}^{i}
%  \end{equation}
% \end{multicols}
%

%
%\begin{equation}
%    s^i \coloneqq h(\Phi(s_{s}^{i}), \Phi(s_{o}^{i})).
%\end{equation}
%The function $h(\cdot, \cdot)$ denotes an aggregation function summarizing the observed topologies of the query entity pair. %, i.e., of the subject and object entity. 
%As aggregation function $h$, we can choose from several alternatives:
%\begin{tabular}[t]{p{4cm}|p{8cm}}
%     $\bullet$~Concat & $h(\Phi(s_{s}^{i}), \Phi(s_{o}^{i})) = \Phi(s_{s}^{i}) || \Phi(s_{o}^{i})$\\
     %$\bullet$\makecell{Element-\\~wise max} & $h_k(\Phi(s_{s}^{i}), \Phi(s_{o}^{i})) = \max(\Phi_k(s_{s}^{i}),  \Phi_k(s_{o}^{i}))$\\
%     $\bullet$~Element-wise max & $h_k(\Phi(s_{s}^{i}), \Phi(s_{o}^{i})) = \max(\Phi_k(s_{s}^{i}),  \Phi_k(s_{o}^{i}))$\\
%     $\bullet$~Sum & $h(\Phi(s_{s}^{i}), \Phi(s_{o}^{i})) = \Phi(s_{s}^{i}) \oplus \Phi(s_{o}^{i})$ \\
%     $\bullet$~Hadamard & $h(\Phi(s_{s}^{i}), \Phi(s_{o}^{i})) = \Phi(s_{s}^{i}) \odot \Phi(s_{o}^{i})$
%\end{tabular}
%where $\Phi_k(\cdot)$ denotes the $k$-th entry of the vector representation.
%In the following, we will use the \emph{Concat} function as aggregation function.

The intuition is that the proposed agent follows links not solely starting at one particular entity, which would result in single inference paths. 
Instead, the state space describes a set of quadruples comprising the local query entities' information.
%Instead, our agent adds all adjacent quadruples of the selected relation type at once to gather a more holistic view helping in succeeding inference steps. 

%%%%%%%%%%%%%%%%%%%
% ACTION SPACE
%%%%%%%%%%%%%%%%%%%
\subsubsection{Action Space $A$}
\label{subsec:action_space}%\hfill\\
Generally, the set of feasible actions is defined as the set of relation types $\mathcal{R}$ within the tKG, i.e., $\mathcal{A} \coloneqq \mathcal{R}$.

At iteration $i$, it comprises all  relation types included in the union of the periphery information of the topologies starting from $u_q$ and $v_q$:
\begin{equation}
    A^i\coloneqq \{p_p | (u_p, r_p, v_p, t_p) \in \cup_{x \in \{u_q,v_q\}} p_{x}^{i} \}
\end{equation}
% \begin{equation}
%     A^i\coloneqq \{\hat{a} | (\hat{u}, \hat{a}, \hat{v}, \hat{t}) \in \cup_{x \in \{sub,obj\}} p_{x}^{i} \}
% \end{equation}

% \begin{equation}
%     A^i \coloneqq rel(p_{sub}^i \cup p_{obj}^i)\,,
% \end{equation}
% %$A^i \coloneqq rel(p_{sub}^i \cup p_{obj}^i)$ 

% included in the union of the periphery information of the topologies starting from $sub$ and $obj$: 
% \begin{equation}
%     A^i \coloneqq rel(p_{sub}^i \cup p_{obj}^i)\,,
% \end{equation}
% %$A^i \coloneqq rel(p_{sub}^i \cup p_{obj}^i)$ 
% where $rel(\cdot)$ extracts the relation types of the \emph{(relation, timestamp)}-pairs.\hfill\\

% The set of feasible actions at iteration $i$ as the relation types included in the union of the periphery information of the topologies starting from $sub$ and $obj$: 
% \begin{equation}
%     A^i \coloneqq rel(p_{sub}^i \cup p_{obj}^i)\,,
% \end{equation}
% %$A^i \coloneqq rel(p_{sub}^i \cup p_{obj}^i)$ 
% where $rel(\cdot)$ extracts the relation types of the \emph{(relation, timestamp)}-pairs.\hfill\\

% \textcolor{red}{[Revise]
% Next, we describe how the agent selects an action $a^i$ at iteration $i$. As described in %section \ref{subsubsec:mira_statespace}
% the previous paragraph, the agent's action space is defined as the union of all \emph{(relation, timestamp)}-pairs being adjacent to $x\text{'s}~\in \{sub,obj\}$ core information. %the core information of either the query subject $s$ or object $o$.
%
%For the description of the action space, we have to consider two aspects. 
%}

\begin{figure*}[t]
    \resizebox{\textwidth}{!}{
        \input{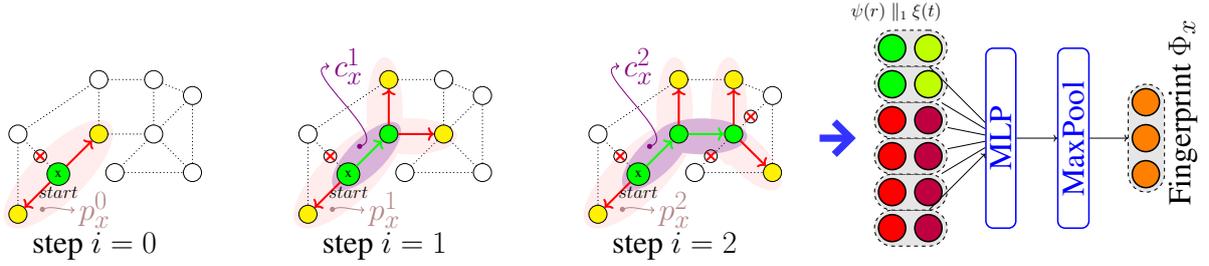}
    }
    \caption{Left: Example of $i=1,2,3$ steps gathering information about the subject/object entity ($x \in \{u_q, v_q\}$). Brighter areas highlight the periphery information $p_x^i$; darker areas highlight the core information $c_x^i$. Red crosses illustrate paths being pruned due to the temporal proximity constrain $tknn=2$; Right: Computation of the graph fingerprint $\Phi_x$; input to the \textit{Multilayer Perceptron} (MLP) constructed by concatenating the embeddings for the relations $\psi(r)$ with their respective temporal embeddings $\xi(t)$ in $c_x$ and $p_x$}
    \label{fig:fingerprint}
\end{figure*}

\begin{example}
Fig. \ref{fig:fingerprint} shows three steps with $tknn=2$ of the agent's %graph 
exploration. For a concise representation, we show the steps only for one query entity $x\in \{u_q,v_q\}$. Starting from $x$, the agent gathers incident relations (cond. i)) fulfilling the temporal proximity (cond. ii)). Red crosses indicate relations being incident to one of the entities within the topology but can be pruned as they are not temporal nearest neighbors. For a query $q_q = (u_q, ?, v_q, t_q)$, the agent chooses amongst $|N_{tknn}(q_q)| = 2$ relations at step $1$. The lighter (darker) areas highlight the periphery (core) information. %, whereas darker areas highlight the core information. 
Initially, the core information is the empty set. The agent's actions are illustrated by green edges. %In each step, the agent expands the topology around the query entity $x$ by relations fulfilling both conditions, and therefore, learns iteratively more about an entity's neighborhood.
Therefore, the agent iteratively learns more about an entity's neighborhood.
\end{example}

%
%%%%%%%%%%%%%%%%%%%
% TRANSITION FNC
%%%%%%%%%%%%%%%%%%%
\subsubsection{Transition}%\hfill\\
The transition function $tr: S \times A \rightarrow S$ is defined as $tr(S^i, a^i) = (c_{u_q}^{i+1}, p_{u_q}^{i+1}, c_{v_q}^{i+1}, p_{v_q}^{i+1})$, where $S^i$ denotes the state in the i-th iteration and $a^i \in A^i$ denotes the action. In the following, the update procedures for the core and periphery information are unfolded in more detail:

\textbf{Core. } Initially, $c_x^0$ is empty. The information stored in $c_x^i$ is extended iteratively with quadruples taken from $p_x^i$ where the relation type matches the agent's chosen action $a^i$ in the $i$-th iteration. We define:
\begin{equation}
\begin{split}
    \label{apptek:eq:core}
    c_x^i \coloneqq \left\{c_{x}^{i-1} \cup y^{i}\right\}  \qquad c_x^0 \coloneqq \varnothing \,, \qquad \text{where}\\
    y^{i} = \{q| \forall q=(u_p, r_p, v_p, t_p) \in p_x^i: r_p = a^i\}.
\end{split}
\end{equation}
Intuitively, $y^i$ denotes the subset of quadruples for which the relation type matches the agent's selected action $a^i$. 
To conclude, $c_x^i$ comprises for each iteration the history of all traversed facts. 

\todo{update of periphery}
\textbf{Periphery. } 
Subsequent to the update procedure of the core's information according to Eq. \ref{apptek:eq:core}, the update of the periphery is initiated. Intuitively, it is updated by quadruples being reachable by the updated core's information such that Eq. \ref{apptek:eq:periphery} holds again.
\textcolor{red}{
%After the agent decides upon which action $a^i$ to explore next, the core information is first updated according to Eq. \ref{eq:core}. Then, $p_x^i$ is extended by %relations 
%quadruples being incident to any entities which are reachable by the information stored in the updated core. We define $p_x^i$ as follows:
% \begin{equation}%
% \label{eq:periphery}%
% \resizebox{\columnwidth}{!}{%
% $p_x^i \coloneqq \{\bigcup_{v \in N_u(a^i)} N(v) \} \cup (p_x^{i-1}\setminus y^{i}) \qquad p_x^0 \coloneqq \{N(x)\}$
% },
% \end{equation}
%where the second term of $p_x^i$ denotes the information from the previous step for which traversed quadruples %pairs 
%(Eq. \ref{eq:core}) are removed. For the first term, entries of $N_u(a^i)$ have the form $(u, a^i, v, t) \subseteq c_x^i$ %$(u, a^i, v, t) \subseteq Core^i$ with $Core^i$ referring to the quadruples being reachable in the core of a topology
%, and $N(v)$ defines the %relation 
%neighborhood of entity $v$, i.e., %\emph{(relation, timestamp)}-pairs incident to $v$
%quadruples with $v$ as the head entity, that are not yet covered by the core information.
}
%
%%%%%%%%%%%%%%%%%%%
% REWARD FNC
%%%%%%%%%%%%%%%%%%%
\subsubsection{Reward $R$}
\label{subsubsec:reward}%\hfill\\
%The agent is expected to learn useful topologies only when they are valuable indicators to predict the correct predicate between the query entities. It
The agent receives a terminal reward of $1$ if the query entities' core information overlap in at least one entity, and $0$ otherwise. All intermediate states receive a reward of $0$. We define the reward as: 
% \begin{equation}
% \label{eq:reward}
%     \mathbbm{1}\{\bigcup_{(r,t) \in c_s^{i+1}} N(r) \cap \bigcup_{(r,t) \in c_o^{i+1}} N(r)\},
% \end{equation}
% \begin{equation}
% \label{eq:reward}
%     r^i=\mathbbm{1}\{ent(Core_{sub}^{i+1}) \cap ent(Core_{obj}^{i+1})\},
% \end{equation}
\begin{equation}
\label{eq:reward}
    \text{reward}^i=\mathbbm{1}\{ent(c_{u_q}^{i+1}) \cap ent(c_{v_q}^{i+1})\},
\end{equation}
where $ent(\cdot)$ extracts the subject and head entities of quadruples. 
%where $N(r)$ yields the set of entities being reachable by following the relation type $r$. 
%
\todo{check}
%Let $Q_\theta (s^i, a^i)$ be the Q-function where $\theta$ denotes the model's parameters. The Q-function defines the long-term reward of taking action $a^i$ at state $s^i$, and following the optimal policy after that. The objective is to learn a value function approximator maximizing Q-values for identifying topologies connecting entities. 

\paragraph{}
% Jede episode beschreibt eine Abfolge von Rels
% Konstruktion des feature vecs

An episode in this MDP ends after receiving a reward or performing a maximum of $I$ steps. Each query $(u_q,?,v_q,t_q)$ might result in a different sequence of action or relation types which can be directly interpreted as reasoning sequence upon the $tKG$. Furthermore, we can apply these reasoning sequences to any query tuple and receive a binary result if the path connects subject and object. Thus, using $m$ reasoning sequences can be used to generate an $m-$dimensional binary feature vector which can be used in the downstream path of predicate prediction. For prediction, we use a linear layer with a successive softmax function for classification.
%To summarize, for predicate prediction, the reasoning about the relation type is based on both topologies. Likewise to \cite{lao-etal-2011-random,xiong-etal-2017-deeppath}, these can be used as logical formulas to infer the relation type. In other words, the topologies are used as binary features (cf. Eq. \ref{eq:reward}) upon which a classifier is learned. In Sec. \ref{subsec:relation_prediction}, we use a simple linear layer for this downstream task.

\subsection{Model architecture}
\label{sec:model_architecture}
%Next, we introduce the architecture of MIRA in more detail. It is 
%Our model, MIRA, is divided into two modules: one module to compute a representation (fingerprint) of the observed topology%, referred to as a \emph{fingerprint}
%; and one to compute Q-values while masking out unreachable actions.

\subsubsection{Calculating Fingerprints}%
\label{subsubsec:fingerprint}%\hfill\\
%
% MAX ARCHITECTURE
%
On traversing the tKG starting from $x \in \{u_q, v_q\}$, the agent iteratively gathers information about their respective neighborhood. As discussed in Sec. \ref{sec:mdp}, the current state $s_x^i$ of $x$ at iteration $i$ is defined by a topology's core information $c_x^{i}$ and its periphery $p_x^{i}$. The computation of the fingerprint translates the herein stored information about the relations and their respective timestamp into a $d_f$-dimensional vector. First of all, we apply an embedding layer $\psi: r \rightarrow \mathbbm{R}^{d_r}$ to represent the relations in $s_x^i$, and correspondingly, an embedding layer $\xi: t \rightarrow \mathbbm{R}^{d_t}$ for the temporal information.
%. Correspondingly, an embedding is applied to encode the time $t$ of relations $r \in x$ denoted by $\xi: t \rightarrow \mathbbm{R}^{d_t}$.
%This is accomplished by concatenating embeddings for relations, denoted by $\psi: r$ \rightarrow \mathbbm{R}^{d_r}$, with their corresponding embeddings of the temporal domain, denoted by $\xi: t \rightarrow \mathbbm{R}^{d_t}$.
Concatenating both embeddings, we receive the following matrix representation of $s_x^i$:
\begin{equation}
    \mathbbm{X}_{s^i_x} = \underset{(u, r, v, t) \in s_x^{i}  }{\parallel_0} (\psi(r)\parallel_1 \xi(t)),
\end{equation}
where $\parallel_l$ denotes the concatenation in the $l$-th dimension. %Therefore, 
Thus, $\mathbbm{X}_{s^i_x}$ has the shape $|s_x^{i}| \times (d_r + d_t)$.
%The input to our state representation function is a $|c_x^{i} \cup p_x^{i}| \times (d_r + d_t)$ dimensional matrix denoted by $\mathbbm{X}_{c_x^{i} \cup p_x^{i}}$, where $d_r$ denotes the dimension of the embedding of the relations, and $d_t$ the dimension of the temporal embeddings. 
Next, $\mathbbm{X}_{s^i_x}$ is fed into an MLP which generates a $|s_x^{i}| \times d_f$-dimensional vector. Finally, a max-pooling operation on the $0$-th dimension yields a $d_f$-dimensional fingerprint. %
%We substitute this series of transformation by $\Phi_x$.
We describe this non-linear transformation by $\Phi : s_x \rightarrow \Phi(s_x)\in \mathbbm{R}^{d_f}$.
Finally, the overall low-dimensional state representation is defined as the  concatenation of the query's subject and object representation:  
\begin{equation}
\tilde{s}^i \coloneqq \Phi(s_{u_q}^{i}) || \Phi(s_{v_q}^{i})
\end{equation}

%
%The fingerprints are calculated for both topologies %starting from $s_q$ and $o_q$ of a query, i.e., we compute
%yielding $\Phi_s$ and $\Phi_o$.
\begin{example}
The calculation of a fingerprint is illustrated in Fig. \ref{fig:fingerprint} (right). The core's information of the topology (green) is concatenated with the relations' information given in the periphery (red). The input to the MLP is provided by the concatenation of the relation embeddings with their respective temporal embeddings. Applying a maxpooling operation, the network yields the fingerprint $\Phi_x$ of the observed topology.
\end{example}

% Afterwards, the state representation (\emph{fingerprint)} is calculated by a non-linear transformation function $\Phi : s_x \rightarrow \Phi(s_x)\in \mathbbm{R}^{d_f}$ being explained in more detail in Sec.  \ref{subsubsec:fingerprint}. Finally, the overall state representation is a concatenation of subject and object representation:  
% \begin{equation}
% \tilde{s}^i \coloneqq \Phi(s_{sub}^{i}) || \Phi(s_{obj}^{i})
% \end{equation}

\subsubsection{Q-network}\label{sec:qlearning}%\hfill\\
The RL frameworks in \cite{xiong-etal-2017-deeppath,das-2017-minerva,lin-2018-multihop,Shen2018MWalkLT} have to handle large action spaces leading to problems whenever we lack training labels for multiple actions. %In the worst case, there could be no samples for certain relations that lead to random outputs. 
Each of the earlier works proposed workarounds either by defining a more sophisticated reward function \cite{xiong-etal-2017-deeppath,lin-2018-multihop} or by combining the learning procedure with MCTS \cite{Shen2018MWalkLT} to tackle the sparse reward setting. Our model uses DQN with action masking for relations having been identified in the periphery as described in Sec. \ref{subsubsec:mira_statespace}.

Having computed the fingerprints %(c.f. sec \ref{subsubsec:fingerprint}) 
for both topologies, we concatenate them $\Phi_{u_q}|| \Phi_{v_q}$ % according to the aggregation function $h(\cdot)$ 
(cf. sec \ref{subsubsec:mira_statespace}) and get a $(2 \cdot d_f)$-dimensional input vector for the Q-net. 
Thereafter, we compute the Q-values for all relation types. An MLP is used as function approximator $\hat{q}_{\theta} = \hat{q}_{\theta} (s, a; \theta)$. %where $\theta$ denotes the network's parameters 
%(c.f. section \ref{subsubsec:deepqlearning}).

\todo{check again}
%Next, we discuss the shrinkage of the agent's search space amongst the possible actions at iteration $i$ by applying action masking. As a large action space might lead to poor convergence, we use this additional masking to disguise all relation types being not reachable in the agent's current state, i.e., there are types of relations that are not incident to any entity in any of the two observed topologies, and additionally, that are not in temporal proximity to the query quadruple. Therefore, they are not relevant for the agent's next choice. 

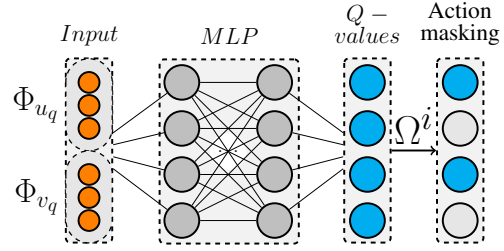
\begin{figure}
\vspace{0pt}%\begin{figure}
    \centering
    \resizebox{0.9\columnwidth}{!}{
        \def\layersep{1.0cm}
\usetikzlibrary{positioning}

\begin{tikzpicture}
	\tikzstyle{rec}=[minimum width=2cm, minimum height=1cm, rounded corners=5mm, dashed, thick, fill=gray!20, draw, anchor=west]
	\tikzstyle{neuron}=[circle, draw, very thick, minimum size=0.75cm,inner sep=0pt, anchor=center]
	\tikzstyle{annot} = [text width=4em, text centered];

	\draw node[minimum width=4cm, minimum height=1cm, rounded corners=1mm, dashed, very thick, draw, anchor=west, fill=gray!10, rotate=90] at (0,-1.5) (input) {};
	
	\draw node[rec, label=north:\huge $\Phi_{u_q}$, rotate=90] at (0,0.5) () {};
	\draw node[rec, label=north:\huge $\Phi_{v_q}$, rotate=90] at (0,-1.5) () {};
    %\draw node[rec, label=north:\huge $r_t^{(i)}$, rotate=90] at (0,-3.5) () {};

    \foreach \color/\name/\y in {orange/-1, orange/-0.5, orange/0, orange/1, orange/1.5, orange/2}
        \draw node[neuron, minimum size=0.45cm, fill=\color] (I-\y) at (0, \y) {};

    % %%% tensor multiplication
    % \tikzset{
    %     cross/.style={path picture={ 
    %     \draw[black, very thick] (path picture bounding box.south east) -- (path picture bounding box.north west) (path picture bounding box.south west) -- (path picture bounding box.north east);
    % }}
    % } 
    % \draw node[circle, cross, draw, very thick, label=below:\huge W, minimum size=7mm] at (12,-0.5) (mult) {};

    %%% stacked action embeddings
    \draw node[minimum width=3cm, minimum height=4cm, rounded corners=1mm, dashed, very thick, draw, anchor=west, fill=gray!10] at (1.5*\layersep,0.5) (MLP) {};
    \foreach \name / \y in {-1, ..., 2}
        \draw node[neuron, fill=lightgray] (H1-\name) at (2*\layersep, \y) {};
    
    \draw node[neuron, fill=white, draw=white] (dots) at (3*\layersep, 0.5) {$\ldots$};
   
    \foreach \name / \y in {-1, ..., 2}
        \draw node[neuron, fill=lightgray] (H2-\name) at (4*\layersep, \y) {};
 
    \draw node[minimum width=4cm, minimum height=1cm, rounded corners=1mm, dashed, very thick, draw, anchor=west, fill=gray!10, rotate=90] at (6*\layersep,-1.5) (pre_out_layer) {};
    \foreach \name / \y in {-1, ..., 2}
        \draw node[neuron, fill=cyan] at (6*\layersep, \y) {};

    \draw node[minimum width=4cm, minimum height=1cm, rounded corners=1mm, dashed, very thick, draw, anchor=west, fill=gray!10, rotate=90] at (8*\layersep,-1.5) (out_layer) {};
    \foreach \color / \name / \y in {gray!20/-1, cyan/0, gray!20/1, cyan/2}
        \draw node[neuron, fill=\color] at (8*\layersep, \y) {};

    % arrows
	%\foreach \source in {-3,...,2}
	\foreach \dest in {-1, ..., 2}
        \path (input) edge (H1-\dest);
    \foreach \src in {-1, ..., 2}
        \foreach \dest in {-1, ..., 2}
            \path (H1-\src) edge (H2-\dest);
	%\foreach \source in {-1,...,2}
    %    \path (H1-\source) edge (dots);
	%\foreach \source in {-1,...,2}
    %    \path (dots) edge (H2-\source);
	 \foreach \dest in {-1, ..., 2}
        \path (H2-\dest) edge (pre_out_layer);

	%%% SoftMax
    %\node[draw, minimum size=1cm, text=blue, rounded corners=2mm, draw=blue, very thick] at (8,0.5) (SoftMax) {\Huge $\hat{q} (s, a)$};
	%\draw[->, very thick] (pre_out_layer) -- (SoftMax);
	
	\draw[->, very thick] (pre_out_layer) -- node[above] {\Huge$\Omega^i$} (out_layer) ;
	\node[annot, above of=pre_out_layer, node distance=2.75cm] () {\Large$Q-values$};
	\node[annot, above of=MLP, node distance=2.5cm] () {\Large $MLP$};
	\node[annot, above of=input, node distance=2.5cm] () {\Large$Input$};
	\node[annot, text width=5em, above of=out_layer, node distance=2.75cm] () {\Large Action masking};

% \node[annot,above of=H-1, node distance=3.5cm] (max_select) {};
\end{tikzpicture}
    }
    \captionof{figure}{Q-Net with action masking}
    \label{fig:qnet}
\end{figure}

%Recall that the agent's possible relations for further exploration are stored within a topology's periphery $p_x^{(i)}$. 
We utilize the information stored in $p_x^{i}$ to mask out
%cloak 
all unreachable semantic links. We define a binary mask $\Omega^i \in \{0,1\}^{|\mathcal{R}|}$ that retains only the information of traversable predicates in the agent's current state. Formally, it is defined as:
\begin{equation}
    \Omega^{i}(r) = 
    \begin{cases} 1 &\text{if } r \in rel(p_{u_q}^{i} \cup p_{v_q}^{i})\\
                0 &\text{otherwise}.
    \end{cases}
\end{equation}
where $rel(\cdot)$ extracts the relation types of quadruples. 
The network's outputs for all predicates is defined by $f (s^i, \theta) \rightarrow [\hat{q}(s^i, a_0), \hat{q}(s^i, a_1), \allowbreak \ldots, \hat{q}(s^i, a_n)]$, where the output $f(s^i, \theta)$ describes the q-values of state $s^i$ under all possible actions in the $i$-th iteration. 
We apply the \emph{Hadamard} product for masking:
\begin{align}
\resizebox{\columnwidth}{!}{
$
    f_{\Omega^i}(s^i, \theta) = [\hat{q}(s^i, a_0), \hat{q}(s^i, a_1), \ldots, \hat{q}(s^i, a_n)] \odot \Omega^i \,,
$   }
\end{align} 
yielding only relevant Q-values the agent has to take into consideration. The architecture is illustrated in Fig. \ref{fig:qnet} where exemplarily two relation types are masked out in the final output.
%
%\begin{minipage}[b]{0.85\columnwidth}
%\end{minipage}
\hfill
%\begin{minipage}[b]{0.85\columnwidth}
% \vspace{0pt}
\begin{table}
\centering
\resizebox{0.9\columnwidth}{!}{
    \begin{tabular}{$c|^c|^c|^c}
\rowstyle{\bfseries}%
          & ICEWS14 & ICEWS05-15 & YAGO15k \\
        \toprule\midrule
         \# Entities    & 7,128  & 10,094 & 15,403\\ 
         \# Relations   & 230   & 251 & 32\\ 
         \# Temporal Info.  & 365   & 4017 & 188\\ 
         $|$train$|$    & 72,826    & 368,962 & 110,441\\
         $|$validation$|$      & 8,817     & 45,858 & 13,792\\
         $|$test$|$     & 8,789     & 46,056 & 13,813\\
         %\#T-intervals             & 2564    & -       & -\\ 
         \bottomrule
    \end{tabular}
}
    \captionof{table}{Dataset statistics}
    \label{tab:dataset_stats}
\end{table}
%\end{minipage}

%
% SECTION : DEEP Q Learning
%
\subsubsection{Deep Q-Learning - Learning Procedure}%
\label{subsubsec:deepqlearning}%\hfill\\
In deep Q-learning, we apply a function approximator $\hat{q}_{\theta} = \hat{q}_{\theta} (s, a; \theta)$, where $\theta$ are the model's parameters. The objective function being optimized is straightforward defined as:
%\begin{equation}
$\mathcal{L}(\theta) = \mathbbm{E}\left[ \left(r(s^i, a^i) + \gamma \max_a \hat{q}_\theta (s^{i+1},a) - \hat{q}_\theta (s^i, a^i) \right)^2 \right]
$
%\end{equation}

\noindent%
For an effective learning procedure, using a \emph{Replay Memory} is of discernible utility \cite{mnih2013atari}. Hence, we store the transitions $(s^i, a^i, \text{reward}^i, s^{i+1})$ in $\mathcal{D}$, and a training batch is generated by sampling from the replay memory, $(s^j, a^j, \text{reward}^j, s^{j+1})\sim \mathcal{D}$. 
\todo{shortened}
%Generally, it is beneficial to learn from the same transitions multiple times which greatly facilitates a stable learning process.
%
%It is generally known, the benefit lies in the fact that this allows for learning from the same transition multiple times and greatly improves a stable learning process.
\section{Evaluation}
\label{sec:evaluation}
\iffalse
In the following, we present results on the \emph{predicate prediction task} (c.f. sec \ref{sec:problem_definition}) and compare them to state-of-the-art embedding based methods on tKGs that have been adapted to our task at hand. Embedding-based methods lack in semantical information of their reasoning. Hence, we highlight our agent's effectiveness w.r.t rewards, how temporal nearest neighbors influence the action space and examine the inferred topologies for explanations of its reasoning.
\fi

\subsection{Experimental Settings}

% \begin{table}
% \centering
% \resizebox{0.4\columnwidth}{!}{
%     \begin{tabular}{$l|^c|^c|^c}
% \rowstyle{\bfseries}%
%           & ICEWS14 & ICEWS05-15 & YAGO15k \\
%         \toprule\midrule
%          \# Entities    & 7128  & 10488 & 15403\\ 
%          \# Relations   & 230   & 251   & 32\\ 
%          \# Temporal Info.  & 365   & 4017 & 188\\ 
%          %\#T-intervals             & 2564    & -       & -\\ 
%          \bottomrule
%     \end{tabular}
% }
%     \caption{Dataset statistics}
%     \label{tab:dataset_stats}
% \end{table}

%\subsubsection{Datasets}\hfill\\
\textbf{Datasets.~}
The \emph{Integrated Crisis Early Warning System} (ICEWS) dataset \cite{garcia-duran-etal-2018-learning} encodes political events with timestamps. 
\todo{commented as footnote provides further info}
%These events interrelate entities (e.g., countries, presidents, etc.) via logical predicates (e.g., \emph{’Make a visit’}, \emph{’Accuse’}, etc.) 
\footnote{Additional information can be found at http://www.icews.com/}. The repository is organized in dumps storing information about events that occurred from 1995 to 2015: 
\emph{ICEWS14} contains events of 2014; and \emph{ICEWS05-15} contains events from 2005 to 2015.

The \emph{YAGO15K} dataset \cite{garcia-duran-etal-2018-learning} is a modification of FB15k \cite{Leblay-ttranse-2018} including either the temporal information \emph{$<$occursSince$>$} or \emph{$<$occursUntil$>$} for only some facts. % Because of its incompleteness regarding the temporal information, we adapt the concatenation step for calculating the fingerprints described in section \ref{subsubsec:fingerprint}.
To capture these scatteredly given temporal information, we modify the shape of the input matrix $\mathbbm{X}_{s_x}$ to a $|s_x| \times (d_r + 2d_t)$ matrix, i.e., we obtain one $d_t$-dimensional vector for \emph{$<$occursSince$>$}, and one for \emph{$<$occursUntil$>$}.

The statistics of the datasets are summarized in Table \ref{tab:dataset_stats}.

%\subsubsection{Adaptions of Competitors}\hfill\\
\noindent%
\textbf{Adaptions of Competitors. }\\
\emph{TNTComplex. }In \cite{Lacroix2020Tensor}, the authors set up an objective function being suitable for entity prediction queries, i.e., $(u_q, r_q, ?, t_q)$.
\todo{commented out Eq.}
%In order to be comparable to our task of
For predicate prediction, %i.e., inferring the relation of a query quadruple, 
we modify the loss function along the relational tube of the 4-order tensor.\\ %(see Appendix \ref{appendix:competitors} for details).
%Note that the formulation of the loss presented in \cite{Lacroix2020Tensor} suits only to queries of the type $(subject, predicate, ?, time)$. 
%As in our work, we focus on the predicate prediction task (i.e., we want to infer the predicate), we slightly modify their loss function. 
%
Following their notation, for each train tuple $(i, j, k, l)$, we define their loss function as:
\begin{equation*}
\resizebox{\columnwidth}{!}{
 $ \Tilde{l}(\hat{X}; (i,j,k,l)) = - \hat{X}_{i,j,k,l} + \log \left(\sum_{j'} exp(\hat{X}_{i,j',k,l})\right)$
}
\end{equation*}

\noindent \emph{DE-TransE/-DistMult/-SimplE. }In \cite{goel2020diachronic}, the authors present a diachronic embedding building on-top of existing embedding methods for static KGs. 
%Following their notation, by including the temporal information for a pair $(v,t)$, where  $v \in \mathcal{V}$ ($\mathcal{V}:$ set of entities), $t \in \mathcal{T}$ ($\mathcal{T}:$ set of timestamps), the method yields a vector $z_v$ in a $d$-dimensional space. For a triple $(v,r,u)$, the scoring function underneath uses the diachronic embedding representation for the various methods.
%For DE-TransE, we get $\phi(v,r,u) = - \|z_v + z_r - z_u\|$, for DE-DistMult $\phi(v,r,u) = \langle z_v, z_r, z_u\rangle$ and for DE-SimplE $\phi(v,r,u) = \left[ \langle z_v, z_r, z_u \rangle + \langle z_u, z_{r^{-1}}, z_v \rangle \right] / 2$. 
\todo{eq commented out}
Due to the associative property of the scoring functions, the objective functions stay the same for the predicate prediction task. %(see Appendix \ref{appendix:competitors} for details). 
However, for evaluation, instead of corrupting the subject/objects entities, we falsify the relations and compute the ranking metrics accordingly. As the temporal information is only sparsely annotated in YAGO15k, the missing information refers to the same (null) embedding vector such that the diachronic embeddings can be at least applied for the existing temporal information.

%The diachronic embedding only applies for timestamps. As in YAGO15K, the temporal information is only sparsely annotated, results are omitted. %and not for intervals. Consequently, results on YAGO15K are omitted.
\begin{table*}[t]
    \centering
    \resizebox{0.99\textwidth}{!}{
%    \begin{tabular}{r| *{3}{R} m{1pt} *{3}{R} m{1pt} *{3}{R}}
    \begin{tabular}{r| ccc m{1pt} ccc m{1pt} ccc}
        $\downarrow$ Model $|$ Dataset $\rightarrow$ & \multicolumn{3}{c}{ICEWS14} &$|$& \multicolumn{3}{c}{ICEWS05-15} &$|$& \multicolumn{3}{c}{YAGO15K}\\ 
        \cmidrule{2-4} \cmidrule{2-4} \cmidrule{6-8} \cmidrule{10-12}
        & \multicolumn{1}{c}{MRR}
        & \multicolumn{1}{c}{Hits@10} 
        %& \multicolumn{1}{c}{Hits@3} 
        & \multicolumn{1}{c}{Hits@1}
        &$|$& \multicolumn{1}{c}{MRR} 
        & \multicolumn{1}{c}{Hits@10} 
        %& \multicolumn{1}{c}{Hits@3} 
        & \multicolumn{1}{c}{Hits@1} 
        &$|$& \multicolumn{1}{c}{MRR} 
        & \multicolumn{1}{c}{Hits@10} 
        %& \multicolumn{1}{c}{Hits@3} 
        & \multicolumn{1}{c}{Hits@1}
        \\
        \cmidrule{1-4} \cmidrule{6-8} \cmidrule{10-12}
        \multicolumn{12}{l}{\bfseries Predicate Prediction with non-static KG embedding techniques}\\
        % CX [$\dagger$] & 45.50 & 69.73 & 33.87 &|& 48.68 & 72.63 & 37.00 \\
        % TA (CX) [$\diamond$]& 40.97 & 63.87 & 29.58 &|& 49.23 & 72.69 & 37.60 \\
        % HyTE [$\star$] & 24.91 & 65.30 & 2.98  &|& 23.73 & 62.76 & 3.11 \\
        % DE-SimplE & 52.60 & 72.50 & 41.80 &|& 51.30 & 74.80 & 39.20 \\
        % TNT-Complex [$\circ$]& 56.72 & 75.40 & 47.04 &|& 60.58 & 78.50 & 51.15 \\
        % TimePlex(base) & 60.25 & 77.05 & 51.29 &|& 63.91 & 81.42 & 54.62 \\
        % TimePlex [$\ddagger$]& 60.40 & 77.11 & 51.50 &|& 63.99 & 81.81 & 54.51 \\
        \toprule
        TNTComplex \cite{Lacroix2020Tensor} & 0.2691 & 0.5266 & 0.1701 &$|$& 0.2398 & 0.4775 & 0.1327 &$|$& \underline{0.5852} & \textbf{0.9699} & \underline{0.3168}\\
        DE-SimplE \cite{goel2020diachronic} & \textbf{0.5074} & \underline{0.7573} & \textbf{0.3882} &$|$& 0.4289 & 0.6551 & 0.3162 &$|$& 0.2749 & 0.4228 & 0.1852\\
        DE-TransE \cite{goel2020diachronic} & 0.2068 & 0.3119 & 0.1364 &$|$& 0.1928 & 0.3486 & 0.1074 &$|$& 0.3906 & 0.4827 & 0.3120\\
        DE-DistMult \cite{goel2020diachronic} & 0.4613 & 0.7370 & 0.3261 &$|$& 0.3197 & 0.5243 & 0.2129 &$|$& 0.2459 & 0.4298 & 0.1457\\
        RE-GCN \cite{li-2021-regcn} & 0.4756 & 0.7524 & \underline{0.3535} & $|$ & \underline{0.4750} & \underline{0.7483} & \textbf{0.3524} &$|$& - & - & -\\
        
        \cmidrule{1-12}%\\
        \multicolumn{12}{l}{\bfseries Predicate Prediction by Non-static KGR w/ RL}\\
        \midrule
        \textbf{APPTeK} & \underline{0.4889} & \textbf{0.7749} & 0.3421 &$|$& 
        %\textbf{0.4600} & \textbf{0.7679} & \underline{0.3150} 
        %\textbf{0.4731} & \textbf{0.8081} & \underline{.3125}
        %\textbf{0.4803} & %\textbf{0.8184} & %\textbf{0.3178}
        
        % topo 50
        \textbf{0.4844} & \textbf{0.8246} & \underline{0.3222}
        &$|$& \textbf{0.6041} & \underline{0.9547} & \textbf{0.3537}\\
        %MIRA* & 0.4185 & 0.6899 & 0.2948 &|& 0.3840 & 0.6644 & 0.2659 \\ 
%        \bottomrule
    \end{tabular}
    }
    \caption{Predicate prediction results %on \textit{ICEWS14}, \textit{ICEWS05-15}, and \textit{YAGO15K}. 
    Best/Second best results are highlighted in bold/underlined.
    % [$\dagger$] \cite{Lacroix2020Tensor};
    % [$\diamond$] \cite{goel2020diachronic}; 
%    [$\star$] \cite{dasgupta-etal-2018-hyte} ; 
%    [$\ddagger$] \cite{jain-etal-2020-temporal}
%    [$\circ$] \cite{Lacroix2020Tensor};
    }
    \label{tab:link_prediction_large}
    %\vspace{-1cm}
\end{table*}

%

%\subsubsection{Evaluation protocol}\hfill\\
\noindent\textbf{Evaluation protocol.~}
We evaluate all methods in the setting of \emph{predicate prediction}, i.e., for queries $(u_q, ?, v_q, t_q)$. %where we know that the entities $subject$ and $object$ interacted at time $t$ with each other, but the specific predicate is missing.
In the test set, we rank the ground-truth relation type $r$ against all other candidates. %relation types. 
The candidate set %of relation types 
is filtered, i.e., the set %candidate relation types 
for $(u, ?, v, t)$ exclude any $r'$ where $(u, r', v, t)$ appears in the train/ val./ test set.

%\subsubsection{Implementation Details}\hfill\\
\noindent\textbf{Implementation Details. }
Our model is implemented in PyTorch, and trained on a single GPU. 
The MLP for calculating the fingerprint uses two hidden layers, each with $16$ neurons. The output (fingerprint) is a $d_f=16$ dimensional vector. We use an embedding size of $d_r=d_t=10$ for the relations and temporal information. The agent's episode length is restricted to $5$, resulting in a maximal path length of $10$. The MLP of the DQN consists of $2$ layers with $16$ neurons in the first hidden layer and $32$ in the second. As weight decay for the fingerprint network and DQN, we use a value of $0.0001$ and a decay factor of $\gamma=0.99$. For restricting the exploration w.r.t the temporal information, we use $tknn$ in the range $[5, 10, 15, 20, 30]$. 
\todo{CHECK - analyse}
%Generally, increasing the temporal $tkNN$ feeds the agent with more information resulting in higher rewards in early stages. However, computing the fingerprints will slow down. 
The buffer of the replay memory is set to $1.000$. We use RMSprop as optimizer with a learning rate of $0.0001$ and a batch size of $64$. %, which increases the variance of the gradient estimates for the policy net. This enhances an additional exploration of the objective function.
%Due to the flawed experiences in early stages in which the agent applies a random exploration heuristic, there is no need to apply a tunnel-vision optimization on the local optimum. Moreover, 
We apply an $epsilon$ greedy strategy with an exponential decay, where $\epsilon_{start} = 1.0$, $\epsilon_{end} = 0.05$ and $\epsilon_{decay} = 0.00001$. Code availability: $<$git-link-on-publication$>$.

%Even though it can slow down the learning due to bouncing updates, it enhances an additional exploration of the objective function.

\subsection{Predicate Prediction}
\label{subsec:relation_prediction}
We use the standard evaluation metrics used in the literature, i.e., mean reciprocal rank (MRR) and HITS@$k$, where $k \in \{1,10\}$. The MRR is the average of the inverse of the mean rank assigned to the true fact over all candidates, whereas HITS@$k$ is a measurement for the percentage a true fact is ranked within the top-$k$ candidate facts. 
As explained in Sec. \ref{subsubsec:reward}, for our reasoning, we use topologies found by the agent as binary features. For each predicate, we sort the the topologies according to the number of times they connected query entities successfully.
We pick the top-$m \in \{15,25,50\}$ most observed topologies for each relation. Taking more topologies into account yields more distinctive feature vectors, e.g., for ICEWS05-15 the best results were obtained by using $m=50$ topologies.
In Table \ref{tab:link_prediction_large}, we summarize the results of the predicate prediction task on the ICEWS14, ICEWS05-15, and YAGO15K dataset. Our approach reaches results which are in line with state-of-the-art results on ICEWS14, and is the best on ICEWS05-15 where we can include more temporal information. On YAGO15K, where time intervals are scatteredly encoded by the relation types $<$\emph{occursSince}$>$ and $<$\emph{occursUntil}$>$, our method beats the modified TNTComplex and DE-xxx approaches for solving predicate prediction w.r.t the MRR score and Hits@1. The evolution unit of RE-GCN expects as input the temporal information which is not ensured to be given on the YAGO15k dataset, hence, the results are omitted.

\subsection{Effectiveness}
%Next, we discuss the size of the agent's possible actions over time considering the temporal proximity constraint. %We show our model's effectiveness w.r.t the agent's rewards gained over time and discuss the success rate for individual relations within the ICEWS dataset. 

\subsubsection{Influence of Parameter $tkNN$}
\label{subsubsec:eval_param_tknn}%\hfill\\
\textbf{On Predicate Prediction. }Table \ref{tab:var_tknn} shows the results on ICEWS14 with varying $tknn \in \{5, 15, 25, 50\}$. Naturally, the running time for each episode and  consequently, for each epoch increases with taking more information into account. We also observe that there is a sweet spot for $tknn$, where results are best. Hence, while increasing $tknn$, the agent is able to gather more information about the neighborhood of the query entities. However, capping the number of quadruples at some point prevents the influence of temporally unconnected events.\\
\noindent\\
\begin{minipage}[t][][t]{\textwidth}
  \begin{minipage}[b]{0.49\columnwidth}
      \begin{minipage}[b]{0.49\textwidth}
          \includegraphics[width=\textwidth]{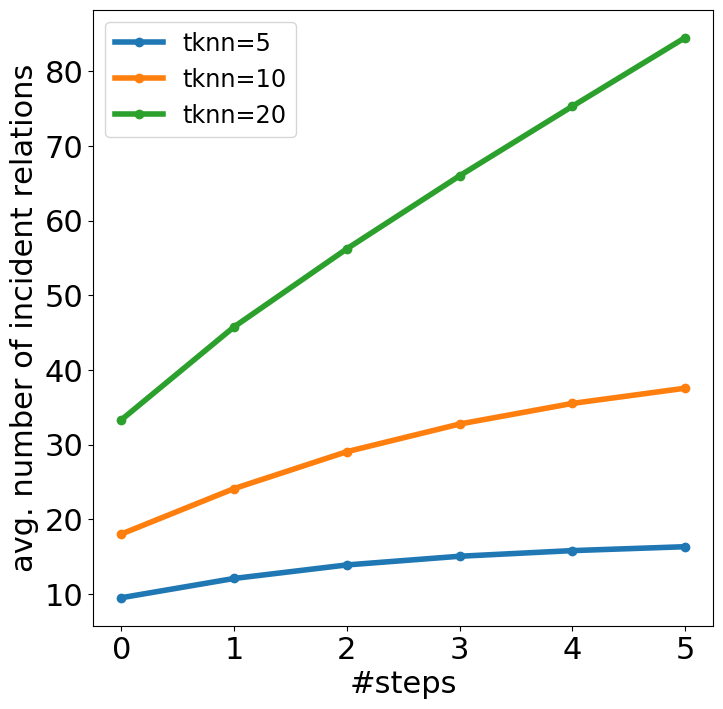}
      \end{minipage}
      \begin{minipage}[b]{0.49\textwidth}
        \includegraphics[width=\textwidth]{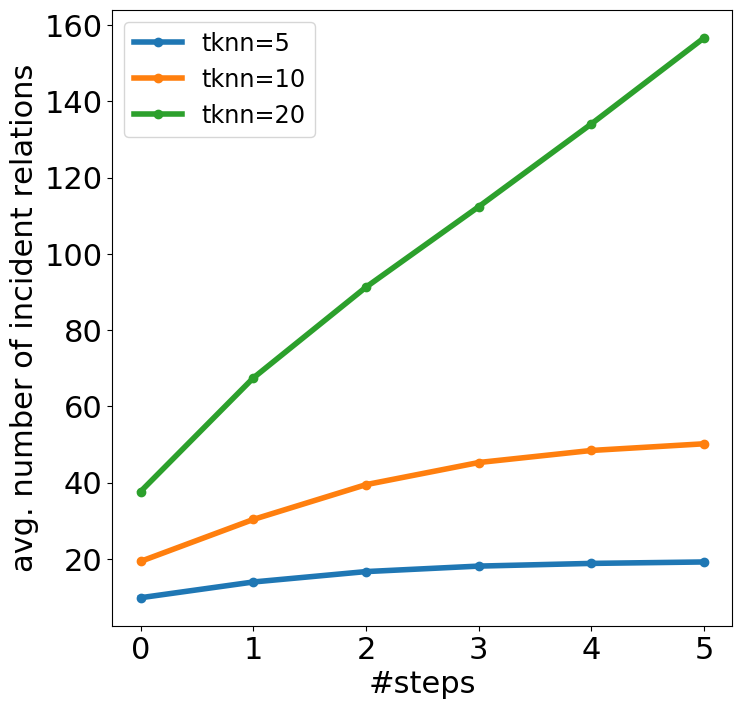}
      \end{minipage}
       \captionof{figure}{Avg. number of possible actions w.r.t the current step for $tknn \in \{5,10,20\}$ on ICEWS14 (left) and ICEWS05-15 (right)}
        \label{fig:tknn}
  \end{minipage}
    \hfill
\end{minipage}
    
% \begin{minipage}[b]{0.49\textwidth}
\begin{table}
  \vspace{0pt}
  \resizebox{\columnwidth}{!}{
    \begin{tabular}[h]{r|c|c|c|c|c}
        tknn & MRR & @10 & @1 & \shortstack{time[ms]/\\episode} & \shortstack{time[s]/\\epoch}\\
        \hline
        \hline
        % HIT@10:  0.7563797209935352
        % HIT@3:   0.4814562776454576
        % HIT@1:   0.28490416241351935
        % MRR:     0.4297885214776804
        5  & 0.4297 & 0.7563 & 0.2849 & \textbf{23} & \textbf{1598}\\
        %
        %
        %HIT@10:  0.7549052965861404    HIT@3:   0.5263695134399455   
        %HIT@1:   0.3183622547351707
        %MRR:     0.4593208931061607   
        15 & 0.4593 & 0.7549 & 0.3183 & 33 & 2245 \\
        %
        %
        %HIT@10:  0.7591017352841103
        %HIT@3:   0.5496200521719405
        %HIT@1:   0.3269819666553249
        %MRR:     0.4716134906779081
        25 & \textbf{0.4716} & \textbf{0.7591} & \textbf{0.3269} & 41 & 2975\\
        %
        %
        %HIT@10:  0.7425428150164455
        %HIT@3:   0.5094703413859589             
        %HIT@1:   0.3079278666212998             
        %MRR:     0.44695625234203096 
        50\vspace{0.5cm} & 0.4469 & 0.7425 & 0.3079 & 66 & 4473 \\
    \end{tabular}
  }
  \captionof{table}{Predicate Prediction on ICEWS14 with varying tkNN, $m=25$, and the respective running times} %per episode, respectively, per epoch}
  \label{tab:var_tknn}
  %\end{minipage}
\end{table}
%\end{minipage}

\noindent\textbf{On Action Space. }Fig. \ref{fig:tknn} (left) illustrates the influence of %the parameter
$tkNN$ on the action space for ICEWS14. It displays the average number of possible relations for $tknn \in \{5,10,20\}$ in each step. While increasing the value for $tknn$, we increase the agent's view on the temporal dimension. Naturally, there is a higher chance for various relation types to be in the candidate set. %relations, and therefore, for different relation types, to be in the candidate set. 
In Fig. \ref{fig:tknn} (right) the effect is illustrated on the larger dataset ICEWS05-15. Due to the more dense nature of the dataset, applying a filtering step on the temporal dimension is of uttermost importance to keep the action space as low as possible, but large enough such that the agent can connect topologies of $u_q$ and $v_q$. %This trade-off is further discussed within the reward history.
%which might occur either less often due to their rareness within the dataset 

\begin{figure}[t]
    \begin{minipage}[c]{0.49\textwidth}
        \begin{subfigure}[b]{.49\textwidth}
            \includegraphics[width=\textwidth]{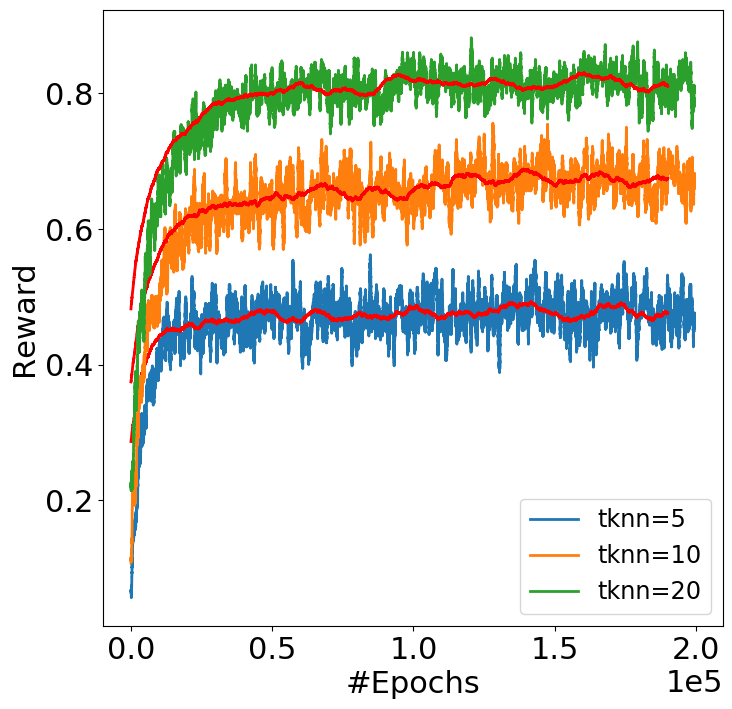}
            \caption{ICEWS14}
            \label{fig:reward_history_icews14}
        \end{subfigure}
        \begin{subfigure}[b]{.49\textwidth}
            \includegraphics[width=\textwidth]{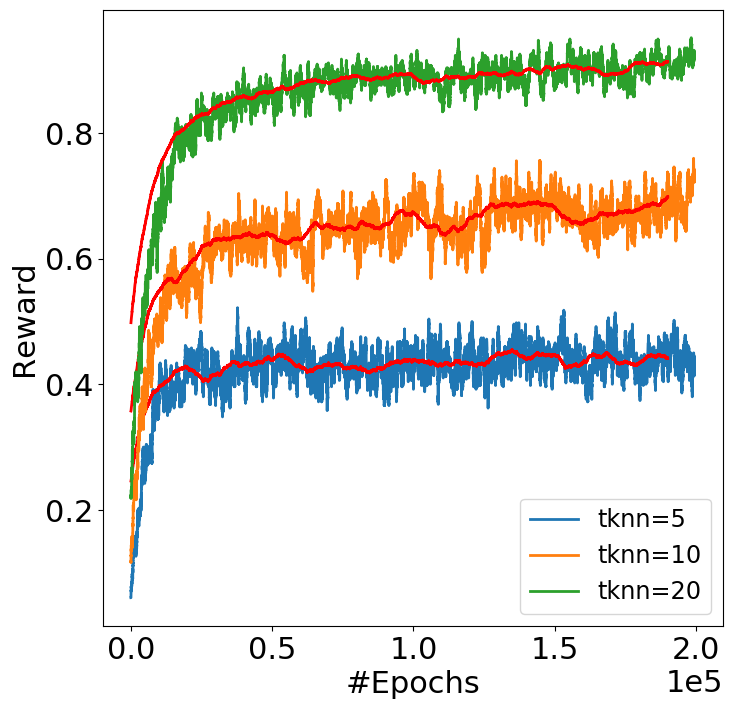}
            \caption{ICEWS05-15}
            \label{fig:reward_history_icews05-15}
        \end{subfigure}        
        %\captionof{figure}{Reward for $tkNN \in \{5,10,20\}$}
        \caption{Reward for $tkNN \in \{5,10,20\}$}
        \label{fig:reward_history}
    \end{minipage}
\end{figure}

\begin{figure}[t]
    \hfill
    \begin{minipage}[c]{0.49\textwidth}
        \begin{subfigure}[b]{.49\textwidth}
            \includegraphics[width=\textwidth]{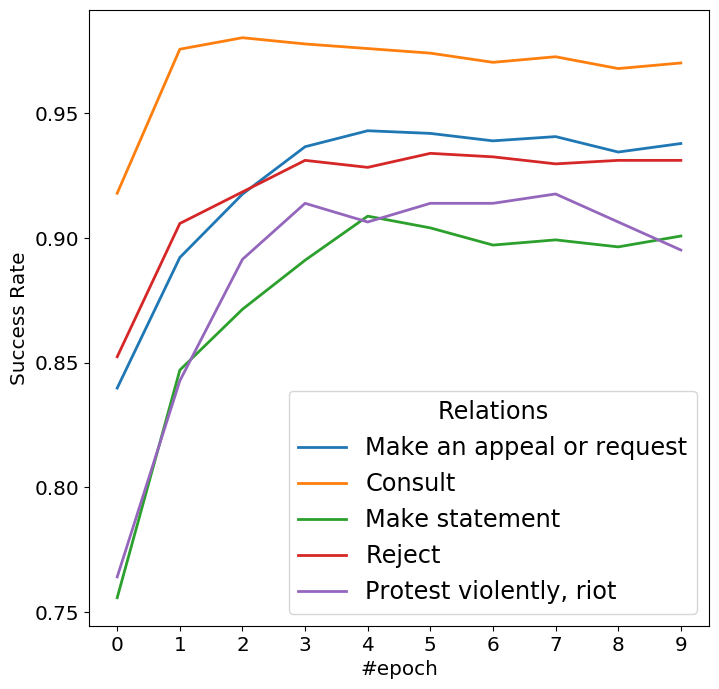}
            \caption{ICEWS14}
            \label{fig:success_rate_icews14}
        \end{subfigure}
        \begin{subfigure}[b]{.49\textwidth}
            \includegraphics[width=\textwidth]{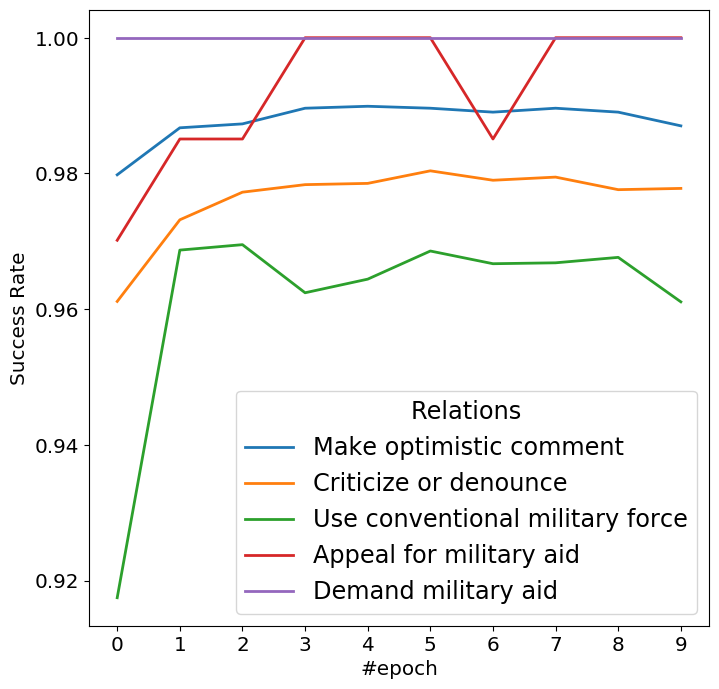}
            \caption{ICEWS05-15}
            \label{fig:success_rate_icews05-15}
        \end{subfigure}
        %\captionof{figure}{Success rate of relation types}
        \caption{Success rate of relation types}
        \label{fig:success_rate}
    \end{minipage}
\end{figure}   

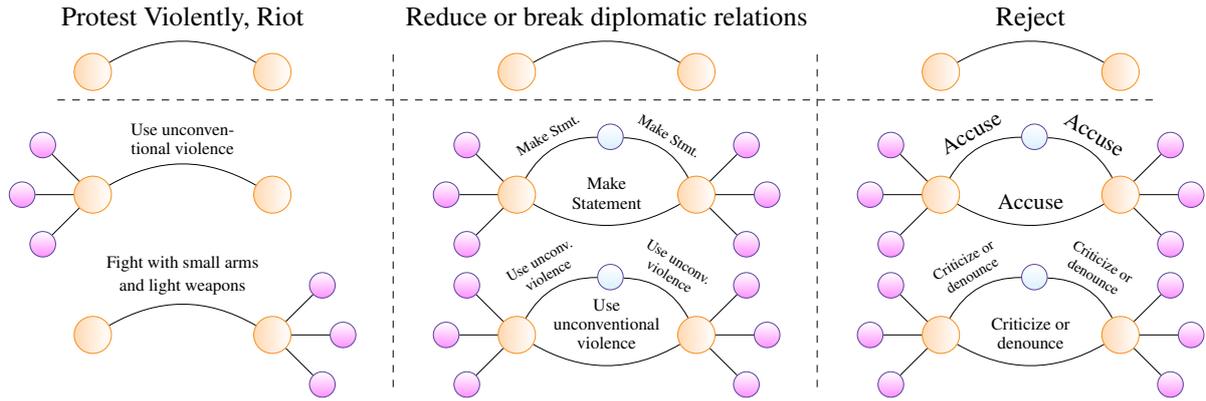
\begin{figure*}[!h]
\centering
\resizebox{1\linewidth}{!}{
\begin{tikzpicture}
\definecolor{lavander}{cmyk}{0,0.48,0,0}
\definecolor{violet}{cmyk}{0.79,0.88,0,0}
\definecolor{burntorange}{cmyk}{0,0.52,1,0}
\definecolor{skyblue}{cmyk}{0.46,0.18,0,0.02}

\def\lav{lavander!90}
\def\oran{orange!30}
\def\sky{skyblue!30}
\tikzstyle{bluepeers}=[draw,circle,violet,bottom color=\sky,
                  top color= white, text=violet,minimum width=10pt]
\tikzstyle{peers}=[draw,circle,violet,bottom color=\lav,
                  top color= white, text=violet,minimum width=10pt]
\tikzstyle{superpeers}=[draw,circle,burntorange, left color=\oran,
                       text=violet,minimum width=15pt]

\draw [dashed] (-0.5, -0.4) -- (15,-0.4);

\begin{scope}
\node[superpeers] (A) {};
\node[superpeers, right = 2cm of A] (B) {};
\path[] (A) edge[bend left] node[above]{Protest Violently, Riot} (B) ;
\end{scope}                  
\begin{scope}[yshift=-1.75cm]
    \node[superpeers] (A) {};
    \node[superpeers, right = 2cm of A] (B) {};
    \foreach \pos/\i in {above left of/1, left of/2, below left of/3}
        \node[peers, \pos = A] (a\i) {};
    \foreach \speer/\peer in {A/a1,A/a2,A/a3}
        \path (\speer) edge (\peer);
    \path[] (A) edge[bend left] node[above]{\scriptsize \shortstack{Use unconven-\\tional violence}} (B) ;
\end{scope}
\begin{scope}[yshift=-3.75cm]
    \node[superpeers] (A) {};
    \node[superpeers, right = 2cm of A] (B) {};
    \foreach \pos/\i in {above right of/1, right of/2, below right of/3}
        \node[peers, \pos = B] (b\i) {};
    \foreach \speer/\peer in {B/b1,B/b2,B/b3}
        \path (\speer) edge (\peer);
    \path[] (A) edge[bend left] node[above]{\scriptsize \shortstack{Fight with small arms\\and light weapons}} (B) ;
\end{scope}

\draw [dashed] (4.25,0.25) -- (4.25,-4.5);

\begin{scope}[xshift=6cm]
\node[superpeers] (A) {};
\node[superpeers, right = 2cm of A] (B) {};
\path[] (A) edge[bend left] node[above]{Reduce or break diplomatic relations} (B) ;
\end{scope}                  
\begin{scope}[xshift=6cm, yshift=-1.75cm]
    \node[superpeers] (A) {};
    \node[superpeers, right = 2cm of A] (B) {};
    \node[bluepeers, above right =0.5cm and 1cm of A] (C) {};
    \foreach \pos/\i in {above left of/1, left of/2, below left of/3}
        \node[peers, \pos = A] (a\i) {};
    \foreach \speer/\peer in {A/a1,A/a2,A/a3}
        \path (\speer) edge (\peer);
    \foreach \pos/\i in {above right of/1, right of/2, below right of/3}
        \node[peers, \pos = B] (b\i) {};
    \foreach \speer/\peer in {B/b1,B/b2,B/b3}
        \path (\speer) edge (\peer);
    \path[] (A) edge[bend right] node[above, yshift=0.1cm]{\scriptsize \shortstack{Make \\Statement}} (B) ;
    \path[] (A) edge[bend left] node[above, midway, sloped]{\tiny Make Stmt.} (C) ;
    \path[] (C) edge[bend left] node[above, midway, sloped]{\tiny Make Stmt.} (B) ;
\end{scope}
\begin{scope}[xshift=6cm, yshift=-3.75cm]
    \node[superpeers] (A) {};
    \node[superpeers, right = 2cm of A] (B) {};
    \node[bluepeers, above right =0.5cm and 1cm of A] (C) {};
    \foreach \pos/\i in {above left of/1, left of/2, below left of/3}
        \node[peers, \pos = A] (a\i) {};
    \foreach \speer/\peer in {A/a1,A/a2,A/a3}
        \path (\speer) edge (\peer);
    \foreach \pos/\i in {above right of/1, right of/2, below right of/3}
        \node[peers, \pos = B] (b\i) {};
    \foreach \speer/\peer in {B/b1,B/b2,B/b3}
        \path (\speer) edge (\peer);
    \path[] (A) edge[bend right] node[above, yshift=0.1cm]{\scriptsize \shortstack{Use \\unconventional \\violence}} (B) ;
    \path[] (A) edge[bend left] node[above, midway, sloped]{\tiny \shortstack{Use unconv. \\violence}} (C) ;
    \path[] (C) edge[bend left] node[above, midway, sloped]{\tiny \shortstack{Use unconv. \\violence}} (B) ;
\end{scope}
    
\draw [dashed] (10.25,0.25) -- (10.25,-4.5);

\begin{scope}[xshift=12cm]
\node[superpeers] (A) {};
\node[superpeers, right = 2cm of A] (B) {};
\path[] (A) edge[bend left] node[above]{Reject} (B) ;
\end{scope}                  
\begin{scope}[xshift=12cm, yshift=-1.75cm]
    \node[superpeers] (A) {};
    \node[superpeers, right = 2cm of A] (B) {};
    \node[bluepeers, above right =0.5cm and 1cm of A] (C) {};
    \foreach \pos/\i in {above left of/1, left of/2, below left of/3}
        \node[peers, \pos = A] (a\i) {};
    \foreach \speer/\peer in {A/a1,A/a2,A/a3}
        \path (\speer) edge (\peer);
    \foreach \pos/\i in {above right of/1, right of/2, below right of/3}
        \node[peers, \pos = B] (b\i) {};
    \foreach \speer/\peer in {B/b1,B/b2,B/b3}
        \path (\speer) edge (\peer);
    \path[] (A) edge[bend right] node[above, yshift=0.1cm]{\footnotesize Accuse} (B) ;
    \path[] (A) edge[bend left] node[above, midway, sloped]{\footnotesize Accuse} (C) ;
    \path[] (C) edge[bend left] node[above, midway, sloped]{\footnotesize Accuse} (B) ;
\end{scope}
\begin{scope}[xshift=12cm, yshift=-3.75cm]
    \node[superpeers] (A) {};
    \node[superpeers, right = 2cm of A] (B) {};
    \node[bluepeers, above right =0.5cm and 1cm of A] (C) {};
    \foreach \pos/\i in {above left of/1, left of/2, below left of/3}
        \node[peers, \pos = A] (a\i) {};
    \foreach \speer/\peer in {A/a1,A/a2,A/a3}
        \path (\speer) edge (\peer);
    \foreach \pos/\i in {above right of/1, right of/2, below right of/3}
        \node[peers, \pos = B] (b\i) {};
    \foreach \speer/\peer in {B/b1,B/b2,B/b3}
        \path (\speer) edge (\peer);
    \path[] (A) edge[bend right] node[above, yshift=0.1cm]{\scriptsize \shortstack{Criticize or\\denounce}} (B) ;
    \path[] (A) edge[bend left] node[above, midway, sloped]{\tiny \shortstack{Criticize or\\denounce}} (C) ;
    \path[] (C) edge[bend left] node[above, midway, sloped]{\tiny \shortstack{Criticize or\\denounce}} (B) ;
\end{scope}
\end{tikzpicture}
}
\caption{ICEWS14: topologies used to infer rel. types (top) of query entities (orange)}
    \label{fig:topologies}
\end{figure*}

\todo{shortened}
\subsubsection{Reward history}
\label{subsubsec:eval_reward_history}%\hfill\\
%As mentioned in Sec. \ref{sec:qlearning}, %and in the previous experiments, 
%one major challenge in reasoning on KG with RL is the large action space. %Compared to recent approaches trying to tackle this issue by steering the agent to sparse rewards, we apply a masking approach. Hence, it disguises relation types which can be disregarded at a particular iteration.
%We apply a masking approach to disguise relation types that can be disregarded at a particular iteration.
%incident to the observed topology at a particular iteration. 
%From the start onwards, the agent is forced to take actions being relevant for the exploration of a topology. %As shown in both illustrations of figure \ref{fig:reward_history}, the agent gains high rewards even after a small number of trained epochs. 
In Fig. \ref{fig:reward_history}, the agent's reward history for %various choices for the parameter 
$tknn \in \{5,10,20\}$ is shown for ICEWS14 (Fig. \ref{fig:reward_history_icews14}) and ICEWS05-15 (Fig. \ref{fig:reward_history_icews05-15}). %Fig. \ref{fig:reward_history_icews14} shows the rewards history on ICEWS14, and fig. \ref{fig:reward_history_icews05-15} the one on ICEWS05-15. 
The colored lines show the reward history with a moving average window of $500$, and the red line with a size of $10.000$.
%whereas the red line shows a smoothing trend with an increased window size of $10.000$.
%
%Even with an increased span of time in the underlying training set, we can find in both settings a convergence to high rewards of $>0.9$ after the first $15.000$ learning steps.
%Note that rewards highly depend on the setting of the temporal nearest neighbors %(c.f. sec \ref{sec:mdp})
%being considered. 
On ICEWS05-15 we reach slightly better rewards for the same values of $tknn$. %A reason for this
This is due to the sparseness of ICEWS14. There are less connections being incident to entities as we only take a time span of 1 year into account. %Generally, there is a trade-off between efficiency and effectiveness. %, in the sense that with higher values of $tkNN$ 
Increasing the value for $tknn$, we consider more relations at each step resulting in more overlapping parts for both topologies. %, and therefore, gather more information for predicting the query relation. 
On the other hand, processing a larger amount of relations increases the running time. %Generally, there is a trade-off between efficiency and effectiveness.
%
%
%With a sufficient density of relations in the dataset, we can shrink the agent's search space. Hence, for the ICEWS05-15, we choose a $tkNN$ of $20$, whereas for the slightly sparser nature of ICEWS14, we increase the neighbors by $tkNN$ of $30$.
%
%This is also accounted for the dense linkage in the ICEWS dataset when we do not shrink down the exploration size by a time interval, i.e., that the relation the agent can take as action has to hold within that time interval, but taking the whole time span of the dataset into account. We will discuss the last issue in more detail in section \ref{sec:eval_qualitative}. In the next section, we will have a closer look into individual links in the ICEWS dataset.
%

%\inlineheader{Success Rate for Individual Relations}
\subsubsection{Success Rate for Individual Links}
\label{subsubsec:eval_succress_rate}%\hfill\\
%This part shows the success rate of individual links. %
%Fig. \ref{fig:success_rate_icews14} illustrates the success rate of $5$ individual links in ICEWS14. %, especially for ones which are predominant in the dataset, for example, \emph{'Make statement'}, and \emph{'Consult'}. 
\todo{shortened}
%In the training procedure, we disregard the relation type of the query. %As can be seen, 
%The agent explores topologies and finds patterns to close a path between query entities. %over time. 
%These patterns are helpful for the inference step of the query relation.
%
%It shows that the agent is able to explore topologies and finds patterns which are helpful for the inference step when disregarding the query relation type in the training procedure. 
%
%Note that this accounts for relation types which are predominant in the dataset, like \emph{'Make statement'} or \emph{'Consult'}, as well as for types occurring less often, e.g.,\emph{'Protest violently, riot'}.
%
%Using our approach shows that the agent is still able to find topologies that help in inferring relations even when we disregard those predominant relations in the training procedure. Even though the disbalance of relation type occurrences in the dataset, the topologies also help in inferring other semantic links like \emph{'Protest violently, riot'} occurring less often. 
\noindent%
Fig.\ref{fig:success_rate_icews14}/\ref{fig:success_rate_icews05-15} illustrates the success rate of $5$ individual links in ICEWS14/ICEWS05-15.
%In Fig. \ref{fig:success_rate_icews05-15}, we picked $5$ relations from ICEWS05-15. 
Generally, taking a larger time span into account helps for finding descriptive topologies. %This is due to a more dense information given in the input resulting in a more precise and unique representation of the fingerprints. 
Moreover, some relations have a high correlation connecting the same entities. In such cases, the agent finds instantly a direct path that interconnects both query entities, given that the relations are also temporal nearest neighbors. From a semantical point of view, e.g., \emph{'Make statement'} or \emph{'Consult'} %or \emph{'Criticize or denounce'} 
are predicates used for a verbal exchange, which do correlate and interconnect the same entities within a time frame. 
%Also, relations like \emph{'Demand military aid'} are often preceded by a generic \emph{'Make statement'} relation. 
For densely connected graph structures, the exploration of topologies will result faster in an overlap in at least one entity such that the success rate reaches a maximum in early stages.

%\begin{figure*}
%\begin{subfigure}[b]{.4\textwidth}
%        \resizebox{\textwidth}{!}{
%            \includegraphics[width=\linewidth]{success_rate_icews14.png}
%        }
%        \caption{ICEWS14}
%        \label{fig:success_rate_icews14}
%    \end{subfigure}
%    %\hfill
%    \begin{subfigure}[b]{.4\textwidth}
%        \resizebox{\textwidth}{!}{
%            \includegraphics[width=\linewidth]{success_rate_icews05-15.png}
%        }
%        \caption{ICEWS05-15}
%        \label{fig:success_rate_icews05-15}
%    \end{subfigure}
%    \caption{Success rate}
%    \label{fig:success_rate}
%\end{figure*}

\newcommand*{\MinNumber}{0.0}%
\newcommand*{\MidNumber}{0.5} %
\newcommand*{\MaxNumber}{1.0}%

%Apply the gradient macro
\newcommand{\ApplyGradient}[1]{%
        \IfDecimal{#1}{
        \ifdim #1 pt > \MidNumber pt
            \pgfmathsetmacro{\PercentColor}{max(min(100.0*(#1 - \MidNumber)/(\MaxNumber-\MidNumber),100.0),0.00)} %
            \hspace{-0.33em}\colorbox{green!\PercentColor!yellow}{#1}
        \else
            \pgfmathsetmacro{\PercentColor}{max(min(100.0*(\MidNumber - #1)/(\MidNumber-\MinNumber),100.0),0.00)} %
            \hspace{-0.33em}\colorbox{red!\PercentColor!yellow}{#1}
        \fi}
        {#1}
}
\newcolumntype{R}{>{\collectcell\ApplyGradient}c<{\endcollectcell}}

\subsection{Qualitative Analysis of Topologies}
\label{sec:eval_qualitative}
%Lastly, we discuss some topologies being used for relation reasoning. 
In Fig. \ref{fig:topologies}, we present exemplary topologies. %, %being explored by the agent starting from the query entities, 
%in order to infer the relation type holding between them. 
%In comparison to embedding methods on tKGs, where we lose track of the explainability of the reasoning, here, we get a better insight into the paths being explored by the agent. 
%The illustration is subdivided into three columns: 
The query relation between two entities (orange) is shown on top of each column; two relevant topologies %having been found by the agent 
are shown below. %the query. 
For predicting, e.g., %the relation type 
\emph{'Protest violently, Riot'}, one way to find a path to the object is by expanding the subject's topology by \emph{'Use unconventional violence'} fulfilling the temporal constraint w.r.t the query. As the agent expands according to a relation type and not by single quadruples, we might get additional connections to other entities (lavender). %Another indicator is given by expanding the object by \emph{'Fight with small arms and light weapons'}. 
%Note that these $1$-hop paths have been found with a temporal proximity to the query timestamp. 
\iffalse
Hence, the model can be used to get tunnel-visioned view w.r.t to the temporal dimension via tuning the $tkNN$ parameter. 
\fi
%Moreover, both expansions are also in semantical accordance with our (human) common-sense of this topic, where weapons might be in use within a violently protest. 
For %the prediction of 
\emph{'Reduce or break diplomatic relations'}, the agent found a connection %from the subject to  object 
by expanding both query entities by %relations being of type 
\emph{'Make Statement'}. As the latter is a predominant and very generic relation type, it occurs more often in the dataset. For relational reasoning, these kind of relation types yield indistinguishable features exacerbating the prediction when used as a single feature. \todo{check}%However, the agent found connections between subject and object either via a direct $1$-hop connection or via a $2$-hop path linkage where an intermediate entity (blue) is used. Whereas in the former case a single expansion for either query entity would be sufficient, we need the expansion on both sides in the latter case. 

%This can also be seen in the
%
%Same applies for the second topology, where the relation type \emph{'Use unconventional violence'} is used to explore both query entities.
%
%Note that this topology is different from the \emph{'Protest Violently, Riot'} inference, where just a single entity (subject) has been expanded. 
In the last example, we see topologies being used %for the inference of
to infer \emph{'Reject'}. %Note that 
Both topologies use generic relation types \emph{'Accuse'} and \emph{'Criticize or Denounce'} which might not be decisive features on their own, but using them together as binary path features yields a more conclusive feature vector. Hence, taking more topologies into account facilitates predicate prediction. %, but increases the run time as for each topology, we have to check whether there is a closed path between the subject and object entity.

\section{Conclusion}
\label{sec:conclusion}
We propose an RL approach for predicate prediction on \emph{temporal Knowledge Graphs} (tKG) suitable for human interpretation.
Given a query \emph{(subject, ?, object, time)}, we train an RL agent %exploring the topology of a tKG starting 
that gathers information from both query entities, simultaneously. The most representative topologies connecting two entities can then be used in downstream tasks like predicting the relation type holding between them. Our architecture computes vector representations (\emph{fingerprints}), of the topologies starting from each of the query entities. Afterwards both fingerprints are concatenated and used to calculate Q-values for relation types (actions) being incident to the observed topologies. To capture the temporal aspect when choosing a relation type, we limit the possible action by temporal proximity to the query. Our evaluation shows \emph{i)} the reasoning topologies can be used as features to infer the predicate between two entities; \emph{ii)} our model allows for modeling time intervals; \emph{iii)} the influence of the temporal proximity to the action space; and \emph{iv)}, we additionally gain a more in-depth insight into the reasons for a prediction by analyzing the inferred topologies.

%\section*{Acknowledgements}

\bibliography{custom}

\begin{thebibliography}{36}
\expandafter\ifx\csname natexlab\endcsname\relax\def\natexlab#1{#1}\fi

\bibitem[{Bai et~al.(2021{\natexlab{a}})Bai, Ma, Zhang, and
  Yu}]{bai-2021-tpmod}
Luyi Bai, Xiangnan Ma, Mingcheng Zhang, and Wenting Yu. 2021{\natexlab{a}}.
\newblock \href {https://doi.org/10.1145/3443687} {Tpmod: A tendency-guided
  prediction model for temporal knowledge graph completion}.
\newblock \emph{ACM Trans. Knowl. Discov. Data}, 15(3).

\bibitem[{Bai et~al.(2021{\natexlab{b}})Bai, Yu, Chen, and Ma}]{bai-2021-tpath}
Luyi Bai, Wenting Yu, Mingzhuo Chen, and Xiangnan Ma. 2021{\natexlab{b}}.
\newblock Multi-hop reasoning over paths in temporal knowledge graphs using
  reinforcement learning.
\newblock \emph{Applied Soft Computing}, 103:107144.

\bibitem[{Bordes et~al.(2013)Bordes, Usunier, Garcia-Dur\'{a}n, Weston, and
  Yakhnenko}]{bordes-2013-transe}
Antoine Bordes, Nicolas Usunier, Alberto Garcia-Dur\'{a}n, Jason Weston, and
  Oksana Yakhnenko. 2013.
\newblock Translating embeddings for modeling multi-relational data.
\newblock In \emph{Proceedings of the 26th International Conference on Neural
  Information Processing Systems - Volume 2 (NIPS)}, NIPS'13, page 2787–2795,
  Red Hook, NY, USA. Curran Associates Inc.

\bibitem[{Chen et~al.(2018)Chen, Xiong, Yan, and
  Wang}]{chen-etal-2018-variational}
Wenhu Chen, Wenhan Xiong, Xifeng Yan, and William~Yang Wang. 2018.
\newblock Variational knowledge graph reasoning.
\newblock In \emph{Proceedings of the 2018 Conference of the North {A}merican
  Chapter of the Association for Computational Linguistics: Human Language
  Technologies, Volume 1 (Long Papers)}, pages 1823--1832, New Orleans,
  Louisiana. Association for Computational Linguistics.

\bibitem[{Das et~al.(2018)Das, Dhuliawala, Zaheer, Vilnis, Durugkar,
  Krishnamurthy, Smola, and McCallum}]{das-2017-minerva}
Rajarshi Das, Shehzaad Dhuliawala, Manzil Zaheer, Luke Vilnis, Ishan Durugkar,
  Akshay Krishnamurthy, Alex Smola, and Andrew McCallum. 2018.
\newblock Go for a walk and arrive at the answer: Reasoning over paths in
  knowledge bases using reinforcement learning.
\newblock In \emph{ICLR}.

\bibitem[{Das et~al.(2017)Das, Neelakantan, Belanger, and
  McCallum}]{das-etal-2017-chains}
Rajarshi Das, Arvind Neelakantan, David Belanger, and Andrew McCallum. 2017.
\newblock Chains of reasoning over entities, relations, and text using
  recurrent neural networks.
\newblock In \emph{Proceedings of the 15th Conference of the {E}uropean Chapter
  of the Association for Computational Linguistics: Volume 1, Long Papers},
  pages 132--141, Valencia, Spain. Association for Computational Linguistics.

\bibitem[{Dasgupta et~al.(2018)Dasgupta, Ray, and
  Talukdar}]{dasgupta-etal-2018-hyte}
Shib~Sankar Dasgupta, Swayambhu~Nath Ray, and Partha Talukdar. 2018.
\newblock {H}y{TE}: Hyperplane-based temporally aware knowledge graph
  embedding.
\newblock In \emph{Proceedings of the 2018 Conference on Empirical Methods in
  Natural Language Processing}, pages 2001--2011, Brussels, Belgium.
  Association for Computational Linguistics.

\bibitem[{Deng et~al.(2020)Deng, Rangwala, and Ning}]{deng-2020}
Songgaojun Deng, Huzefa Rangwala, and Yue Ning. 2020.
\newblock Dynamic knowledge graph based multi-event forecasting.
\newblock In \emph{Proceedings of the 26th ACM SIGKDD International Conference
  on Knowledge Discovery \& Data Mining}, KDD '20, page 1585–1595, New York,
  NY, USA. Association for Computing Machinery.

\bibitem[{Fu et~al.(2019)Fu, Chen, Qu, Jin, and Ren}]{fu-2019-collaborative}
Cong Fu, Tong Chen, Meng Qu, Woojeong Jin, and Xiang Ren. 2019.
\newblock Collaborative policy learning for open knowledge graph reasoning.
\newblock \emph{EMNLP}.

\bibitem[{Garc{\'\i}a-Dur{\'a}n et~al.(2018)Garc{\'\i}a-Dur{\'a}n,
  Duman{\v{c}}i{\'c}, and Niepert}]{garcia-duran-etal-2018-learning}
Alberto Garc{\'\i}a-Dur{\'a}n, Sebastijan Duman{\v{c}}i{\'c}, and Mathias
  Niepert. 2018.
\newblock Learning sequence encoders for temporal knowledge graph completion.
\newblock In \emph{Proceedings of the 2018 Conference on Empirical Methods in
  Natural Language Processing}, pages 4816--4821, Brussels, Belgium.
  Association for Computational Linguistics.

\bibitem[{Goel et~al.(2020)Goel, Kazemi, Brubaker, and
  Poupart}]{goel2020diachronic}
Rishab Goel, Mehran~Seyed Kazemi, Marcus Brubaker, and Pascal Poupart. 2020.
\newblock Diachronic embedding for temporal knowledge graph completion.
\newblock \emph{National Conference on Artificial Intelligence (AAAI)}.

\bibitem[{Guo et~al.(2019)Guo, Zhang, Hu, Sun, and Qu}]{Lingbing-2019}
Lingbing Guo, Qingheng Zhang, Wei Hu, Zequn Sun, and Yuzhong Qu. 2019.
\newblock {Learning to Complete Knowledge Graphs with Deep Sequential Models}.
\newblock \emph{Data Intelligence}, 1(3):289--308.

\bibitem[{Han et~al.(2020)Han, Ma, Wang, G\"{u}nnemann, and
  Tresp}]{Han2020graph}
Zhen Han, Yunpu Ma, Yuyi Wang, Stephan G\"{u}nnemann, and Volker Tresp. 2020.
\newblock {Graph Hawkes Neural Network for Forecasting on Temporal Knowledge
  Graphs}.
\newblock In \emph{{8th Automated Knowledge Base Construction (AKBC)}}.

\bibitem[{Jin et~al.(2020)Jin, Qu, Jin, and Ren}]{jin2020Renet}
Woojeong Jin, Meng Qu, Xisen Jin, and Xiang Ren. 2020.
\newblock Recurrent event network: Autoregressive structure inference over
  temporal knowledge graphs.
\newblock In \emph{The 2020 Conference on Empirical Methods in Natural Language
  Processing (EMNLP)}.

\bibitem[{Jung et~al.(2021)Jung, Jung, and Kang}]{jung-2021}
Jaehun Jung, Jinhong Jung, and U~Kang. 2021.
\newblock Learning to walk across time for interpretable temporal knowledge
  graph completion.
\newblock In \emph{Proceedings of the 27th ACM SIGKDD Conference on Knowledge
  Discovery \& Data Mining}, KDD '21, page 786–795, New York, NY, USA.
  Association for Computing Machinery.

\bibitem[{Kazemi and Poole(2018)}]{kazemi-2018-simple}
Seyed~Mehran Kazemi and David Poole. 2018.
\newblock Simple embedding for link prediction in knowledge graphs.
\newblock In \emph{Advances in Neural Information Processing Systems},
  volume~31, pages 4284--4295. Curran Associates, Inc.

\bibitem[{Lacroix et~al.(2020)Lacroix, Obozinski, and
  Usunier}]{Lacroix2020Tensor}
Timothée Lacroix, Guillaume Obozinski, and Nicolas Usunier. 2020.
\newblock Tensor decompositions for temporal knowledge base completion.
\newblock In \emph{International Conference on Learning Representations}.

\bibitem[{Lao et~al.(2011)Lao, Mitchell, and Cohen}]{lao-etal-2011-random}
Ni~Lao, Tom Mitchell, and William~W. Cohen. 2011.
\newblock Random walk inference and learning in a large scale knowledge base.
\newblock In \emph{Proceedings of the 2011 Conference on Empirical Methods in
  Natural Language Processing}, pages 529--539, Edinburgh, Scotland, UK.
  Association for Computational Linguistics.

\bibitem[{Leblay and Chekol(2018)}]{Leblay-ttranse-2018}
Julien Leblay and Melisachew~Wudage Chekol. 2018.
\newblock Deriving validity time in knowledge graph.
\newblock In \emph{Companion Proceedings of the The Web Conference 2018}, WWW
  '18, page 1771–1776, Republic and Canton of Geneva, CHE. International
  World Wide Web Conferences Steering Committee.

\bibitem[{Li et~al.(2021{\natexlab{a}})Li, Wang, Rong, and
  Mao}]{Li-2021-MemoryPathAD}
Shuangyin Li, Heng Wang, Pan Rong, and Mingzhi Mao. 2021{\natexlab{a}}.
\newblock Memorypath: A deep reinforcement learning framework for incorporating
  memory component into knowledge graph reasoning.
\newblock \emph{Neurocomputing}, 419:273--286.

\bibitem[{Li et~al.(2021{\natexlab{b}})Li, Jin, Guan, Li, Guo, Wang, and
  Cheng}]{li-2021-search}
Zixuan Li, Xiaolong Jin, Saiping Guan, Wei Li, Jiafeng Guo, Yuanzhuo Wang, and
  Xueqi Cheng. 2021{\natexlab{b}}.
\newblock \href {http://arxiv.org/abs/2106.00327} {Search from history and
  reason for future: Two-stage reasoning on temporal knowledge graphs}.

\bibitem[{Li et~al.(2021{\natexlab{c}})Li, Jin, Li, Guan, Guo, Shen, Wang, and
  Cheng}]{li-2021-regcn}
Zixuan Li, Xiaolong Jin, Wei Li, Saiping Guan, Jiafeng Guo, Huawei Shen,
  Yuanzhuo Wang, and Xueqi Cheng. 2021{\natexlab{c}}.
\newblock \href {http://arxiv.org/abs/2104.10353} {Temporal knowledge graph
  reasoning based on evolutional representation learning}.
\newblock \emph{CoRR}, abs/2104.10353.

\bibitem[{Lin et~al.(2018)Lin, Socher, and Xiong}]{lin-2018-multihop}
Xi~Victoria Lin, Richard Socher, and Caiming Xiong. 2018.
\newblock Multi-hop knowledge graph reasoning with reward shaping.
\newblock In \emph{Proceedings of the 2018 Conference on Empirical Methods in
  Natural Language Processing}, pages 3243--3253, Brussels, Belgium.
  Association for Computational Linguistics.

\bibitem[{Mnih et~al.(2013)Mnih, Kavukcuoglu, Silver, Graves, Antonoglou,
  Wierstra, and Riedmiller}]{mnih2013atari}
Volodymyr Mnih, Koray Kavukcuoglu, David Silver, Alex Graves, Ioannis
  Antonoglou, Daan Wierstra, and Martin Riedmiller. 2013.
\newblock Playing atari with deep reinforcement learning.
\newblock In \emph{NIPS Deep Learning Workshop}.

\bibitem[{Onuki et~al.(2019)Onuki, Murata, Nukui, Inagi, Qiu, Watanabe, and
  Okamoto}]{onuki-2019-jour}
Yohei Onuki, Tsuyoshi Murata, Shun Nukui, Seiya Inagi, Xule Qiu, Masao
  Watanabe, and Hiroshi Okamoto. 2019.
\newblock Relation prediction in knowledge graph by multi-label deep neural
  network.
\newblock \emph{Applied Network Science}, 4(1):20.

\bibitem[{Shen et~al.(2018)Shen, Chen, Huang, Guo, and Gao}]{Shen2018MWalkLT}
Y.~Shen, Jianshu Chen, Po-Sen Huang, Yuqing Guo, and Jianfeng Gao. 2018.
\newblock M-walk: Learning to walk over graphs using monte carlo tree search.
\newblock In \emph{Conference on Neural Information Processing Systems
  (NeurIPS)}.

\bibitem[{Sun et~al.(2021)Sun, Zhong, Ma, Han, and He}]{sun-2021-timetraveler}
Haohai Sun, Jialun Zhong, Yunpu Ma, Zhen Han, and Kun He. 2021.
\newblock Timetraveler: Reinforcement learning for temporal knowledge graph
  forecasting.
\newblock In \emph{Conference on Empirical Methods in Natural Language
  Processing (EMNLP)}.

\bibitem[{Teru et~al.(2020)Teru, Denis, and Hamilton}]{teru-2020-pmlr}
Komal Teru, Etienne Denis, and Will Hamilton. 2020.
\newblock Inductive relation prediction by subgraph reasoning.
\newblock In \emph{Proceedings of the 37th International Conference on Machine
  Learning}, volume 119 of \emph{Proceedings of Machine Learning Research},
  pages 9448--9457. PMLR.

\bibitem[{Trivedi et~al.(2017)Trivedi, Dai, Wang, and
  Song}]{trivedi-2017-knowevolve}
Rakshit Trivedi, Hanjun Dai, Yichen Wang, and Le~Song. 2017.
\newblock Know-evolve: Deep temporal reasoning for dynamic knowledge graphs.
\newblock In \emph{Proceedings of the 34th International Conference on Machine
  Learning (ICML)}, volume~70 of \emph{Proceedings of Machine Learning
  Research}, pages 3462--3471, International Convention Centre, Sydney,
  Australia. PMLR.

\bibitem[{Wang et~al.(2014)Wang, Mazaitis, and Cohen}]{Wang-2014-proppr}
William~Yang Wang, Kathryn Mazaitis, and William~W. Cohen. 2014.
\newblock Structure learning via parameter learning.
\newblock In \emph{Proceedings of the 23rd ACM International Conference on
  Conference on Information and Knowledge Management}, CIKM '14, page
  1199–1208, New York, NY, USA. Association for Computing Machinery.

\bibitem[{Wu et~al.(2020)Wu, Cao, Cheung, and Hamilton}]{wu-2020-temp}
Jiapeng Wu, Meng Cao, Jackie Chi~Kit Cheung, and William~L. Hamilton. 2020.
\newblock {T}e{MP}: Temporal message passing for temporal knowledge graph
  completion.
\newblock In \emph{Proceedings of the 2020 Conference on Empirical Methods in
  Natural Language Processing (EMNLP)}, pages 5730--5746, Online. Association
  for Computational Linguistics.

\bibitem[{Xiong et~al.(2017)Xiong, Hoang, and Wang}]{xiong-etal-2017-deeppath}
Wenhan Xiong, Thien Hoang, and William~Yang Wang. 2017.
\newblock {D}eep{P}ath: A reinforcement learning method for knowledge graph
  reasoning.
\newblock In \emph{Proceedings of the 2017 Conference on Empirical Methods in
  Natural Language Processing}, pages 564--573, Copenhagen, Denmark.
  Association for Computational Linguistics.

\bibitem[{Xu et~al.(2021)Xu, Chen, Nayyeri, and
  Lehmann}]{xu-etal-2021-temporal}
Chengjin Xu, Yung-Yu Chen, Mojtaba Nayyeri, and Jens Lehmann. 2021.
\newblock Temporal knowledge graph completion using a linear temporal
  regularizer and multivector embeddings.
\newblock In \emph{Proceedings of the 2021 Conference of the North American
  Chapter of the Association for Computational Linguistics: Human Language
  Technologies}, pages 2569--2578, Online. Association for Computational
  Linguistics.

\bibitem[{Xu et~al.(2020)Xu, Nayyeri, Alkhoury, Yazdi, and Lehmann}]{Xu-2020}
Chenjin Xu, Mojtaba Nayyeri, Fouad Alkhoury, Hamed~Shariat Yazdi, and Jens
  Lehmann. 2020.
\newblock Temporal knowledge graph completion based on time series gaussian
  embedding.
\newblock In \emph{The Semantic Web - ISWC 2020 - 19th International Semantic
  Web Conference, Athens, Greece, November 2-6, 2020, Proceedings, Part I},
  volume 12506 of \emph{Lecture Notes in Computer Science}, pages 654--671.
  Springer.

\bibitem[{Yang et~al.(2015)Yang, Yih, He, Gao, and Deng}]{yang-2015-embedding}
Bishan Yang, Scott Wen-tau Yih, Xiaodong He, Jianfeng Gao, and Li~Deng. 2015.
\newblock Embedding entities and relations for learning and inference in
  knowledge bases.
\newblock In \emph{Proceedings of the International Conference on Learning
  Representations (ICLR) 2015}.

\bibitem[{Zhu et~al.(2021)Zhu, Chen, Fan, Cheng, and Zhang}]{zhu-2021-cygnet}
Cunchao Zhu, Muhao Chen, Changjun Fan, Guangquan Cheng, and Yan Zhang. 2021.
\newblock Learning from history: Modeling temporal knowledge graphs with
  sequential copy-generation networks.
\newblock In \emph{The 35th AAAI Conference on Artificial Intelligence}.

\end{thebibliography}

\appendix

% \section{Example Appendix}
% \label{sec:appendix}

% This is an appendix.

\end{document}